\PassOptionsToPackage{table}{xcolor}
\documentclass[11pt]{article}

\usepackage{fullpage}
\usepackage[T1]{fontenc}
\usepackage[utf8]{inputenc}
\usepackage{fix-cm}
\usepackage{amsmath}
\usepackage{amsthm}
\usepackage{amssymb}
\usepackage{bm}
\usepackage{graphicx}
\usepackage{booktabs} 
\usepackage{caption}
\usepackage{subcaption}
\usepackage{comment}
\usepackage{microtype} 
\usepackage{verbatim}
\usepackage[square,sort,comma,numbers]{natbib}
\usepackage{CJKutf8}
\usepackage{graphics}
\usepackage{booktabs}
\usepackage{algpseudocode}
\usepackage{algorithm}
\usepackage{enumerate}
\usepackage[dvipsnames]{xcolor}
\definecolor{Gray}{gray}{0.95}
\usepackage[align=center,shadow=true,shadowsize=6pt,nobreak=true,framemethod=tikz,skipabove=9.5pt,skipbelow=9pt,innertopmargin=5pt,innerbottommargin=5pt,innerleftmargin=5pt,innerrightmargin=5pt,leftmargin=1.5pt,rightmargin=1.5pt]{mdframed}
\usetikzlibrary{shadows}
\usepackage{tcolorbox}
\tcbuselibrary{breakable} 

\makeatletter
\numberwithin{equation}{section}
\theoremstyle{plain}

\theoremstyle{definition}

\theoremstyle{plain}

\theoremstyle{remark}
\newtheorem*{rem*}{\protect\remarkname}

\theoremstyle{plain}

\@ifundefined{date}{}{\date{}}
\usepackage{color}
\usepackage{colortbl}
\usepackage{svg}
\usepackage[lined,boxed,ruled, linesnumbered, algo2e]{algorithm2e}

\usepackage[tmargin=.9in, lmargin=.8in, rmargin=.8in, bmargin=.9in]
           {geometry}

\usepackage{amsfonts}
\usepackage{tikz}
\usepackage{enumitem}
\usepackage{wrapfig}
\allowdisplaybreaks

\usepackage{multicol}

\usepackage{xcolor}

\usepackage{amsmath,amsfonts,bm}

\def\eqref#1{equation~\ref{#1}}

\def\1{\bm{1}}

\DeclareMathAlphabet{\mathsfit}{\encodingdefault}{\sfdefault}{m}{sl}
\SetMathAlphabet{\mathsfit}{bold}{\encodingdefault}{\sfdefault}{bx}{n}

\usepackage{svg}
\usepackage{url}
\usepackage{graphicx}
\usepackage{colortbl}
\usepackage{longtable}
\usepackage{array}
\usepackage{booktabs}
\usepackage{wrapfig}
\usepackage{mdframed,lipsum}
\usepackage{multirow}
\usepackage{bbding}
\definecolor{darkblue}{RGB}{25, 50, 112}
\definecolor{c1}{HTML}{fdb4ae}
\definecolor{c2}{HTML}{731d1e}
\definecolor{c3}{HTML}{508AB2}
\definecolor{c4}{HTML}{BFF6BA}

\usepackage[pagebackref=true,breaklinks=true,colorlinks=true,bookmarks=false]{hyperref}
\definecolor{deepred}{HTML}{940000}
\definecolor{green2}{HTML}{BFF6BA}
\hypersetup{linkcolor=deepred}
\hypersetup{urlcolor  = [rgb]{0.4,0.15,0.95}}
\hypersetup{citecolor=[rgb]{0.4,0.15,0.95}}
\usepackage{tabularx} %
\usepackage{listings} %
\usepackage{cleveref}

\usepackage{pifont}
\usepackage{tocloft}
\usepackage[toc,page,header]{appendix}
\usepackage{adjustbox}
\usepackage{minitoc}

\renewcommand \thepart{}
\renewcommand \partname{}

\definecolor{keywordcolor}{rgb}{0.7, 0.1, 0.1}   %
\definecolor{tacticcolor}{rgb}{0.0, 0.1, 0.6}    %
\definecolor{commentcolor}{rgb}{0.4, 0.4, 0.4}   %
\definecolor{symbolcolor}{rgb}{0.0, 0.1, 0.6}    %
\definecolor{sortcolor}{rgb}{0.1, 0.5, 0.1}      %
\definecolor{attributecolor}{rgb}{0.7, 0.1, 0.1} %
\definecolor{darkspringgreen}{rgb}{0.09, 0.45, 0.27}
\usepackage{listings}

\lstnewenvironment{pythoncode}[1][]{
  \lstset{
    language=Python,
    basicstyle=\small\ttfamily,
    numbers=left,
    numberstyle=\tiny,
    stepnumber=1,
    numbersep=5pt,
    frame=single,
    showstringspaces=false,
    keywordstyle=\color{blue},
    commentstyle=\color{gray},
    stringstyle=\color{purple},
    captionpos=b,
    breaklines=true,
    breakatwhitespace=true,
    tabsize=4,
    caption=#1
  }
}{}
\newcommand{\ie}{\emph{i.e.}}
\newcommand{\eg}{\emph{e.g.}}
\newcommand{\our}{FormalMATH}

\newcommand\blfootnote[1]{%
  \begingroup
  \renewcommand\thefootnote{}\footnote{#1}%
  \addtocounter{footnote}{-1}%
  \endgroup
}

\newlength\savewidth\newcommand\shline{\noalign{\global\savewidth\arrayrulewidth
  \global\arrayrulewidth 1pt}\hline\noalign{\global\arrayrulewidth\savewidth}}

\usepackage{fancyhdr}

\usepackage[textsize=tiny,textwidth=0.7in]{todonotes}

\begin{document}
\newcommand{\wy}[1]{\textcolor{cyan}{[Weiyang: #1]}}
\newcommand{\zl}[1]{\textcolor{red}{[Zhouliang: #1]}}
\newcommand{\fix}{\marginpar{FIX}}
\newcommand{\new}{\marginpar{NEW}}
\begin{CJK}{UTF8}{gbsn}

\newcommand{\github}{\raisebox{-1.5pt}{\includegraphics[height=.9em]{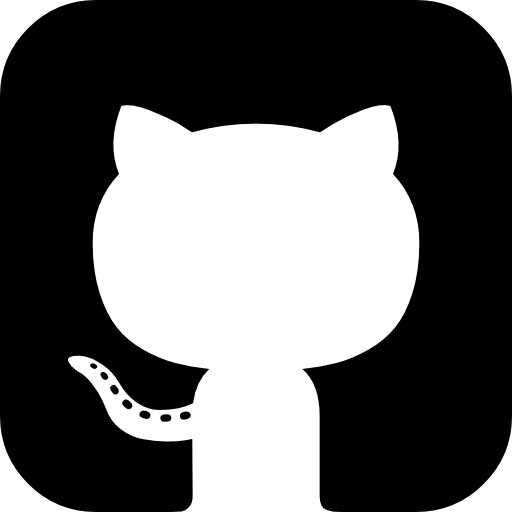}}\xspace}
\newcommand{\projectpage}{\raisebox{-1.5pt}{\includegraphics[height=.9em]{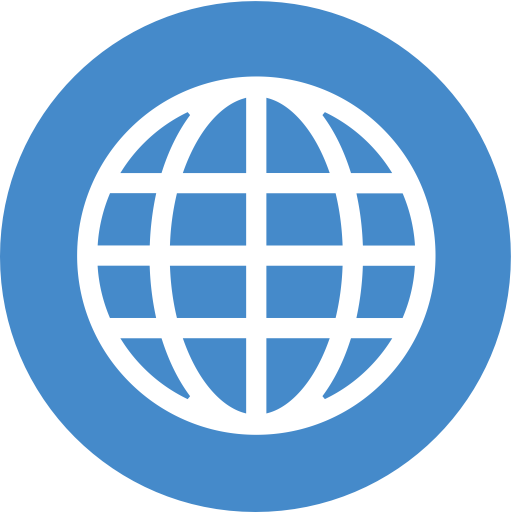}}\xspace}
\newcommand{\dataset}{\raisebox{-1.5pt}{\includegraphics[height=.9em]{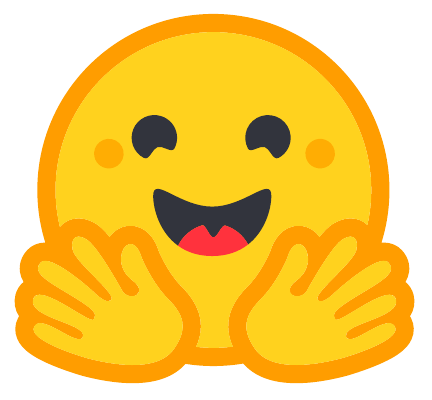}}\xspace}

\title{\vspace{-5mm}FormalMATH: Benchmarking Formal Mathematical\\Reasoning of Large Language Models\vspace{5mm}}

\author{\fontsize{10pt}{\baselineskip}\selectfont
Zhouliang Yu\textsuperscript{1,2,*}, Ruotian Peng\textsuperscript{3,*}, Keyi Ding\textsuperscript{4,*}, Yizhe Li\textsuperscript{5}, Zhongyuan Peng\textsuperscript{4},
Minghao Liu\textsuperscript{5},\\\fontsize{10pt}{\baselineskip}\selectfont Yifan Zhang\textsuperscript{6}, Zheng Yuan\textsuperscript{4}, Huajian Xin\textsuperscript{4}, Wenhao Huang\textsuperscript{4}, Yandong Wen\textsuperscript{3}, Ge Zhang\textsuperscript{4}, Weiyang Liu\textsuperscript{7,\textdagger}\\[1.5mm]\fontsize{10pt}{\baselineskip}\selectfont
\textsuperscript{1}The Chinese University of Hong Kong~~~\textsuperscript{2}Numina~~~\textsuperscript{3}Westlake University~~~\textsuperscript{4}M-A-P\\[0.25mm]\fontsize{10pt}{\baselineskip}\selectfont
\textsuperscript{5}2077AI~~~\textsuperscript{6}University of California, Los Angeles~~~\textsuperscript{7}Max Planck Institute for Intelligent Systems, T\"ubingen\vspace{-6.5mm}
}

\maketitle
\doparttoc 
\faketableofcontents

\thispagestyle{fancy}
\fancyhf{}
\fancyhead[R]{\vspace{-17mm}\includegraphics[height=.58cm]{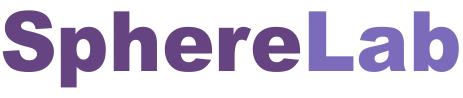}\hspace{7mm}\includegraphics[height=.67cm]{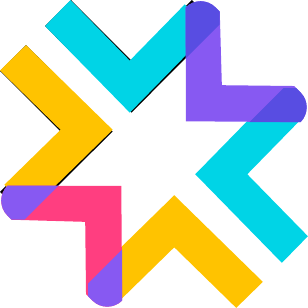}\includegraphics[height=.6cm]{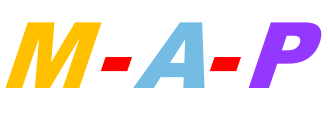}\hspace{-5mm}}
\renewcommand{\headrulewidth}{0pt}

\blfootnote{Technical report v1~~~\textsuperscript{*}Equal  contributions~~~\textsuperscript{\textdagger}Corresponding author
}

\begin{abstract}
Formal mathematical reasoning remains a critical challenge for artificial intelligence, hindered by limitations of existing benchmarks in scope and scale.
To address this, we present FormalMATH, a large-scale Lean4 benchmark comprising 5,560 formally verified problems spanning from high-school Olympiad challenges to undergraduate-level theorems across diverse domains (\eg, algebra, applied mathematics, calculus, number theory, and discrete mathematics).
To mitigate the inefficiency of manual formalization, we introduce a novel human-in-the-loop autoformalization pipeline that integrates: (1) specialized large language models (LLMs) for statement autoformalization, (2) multi-LLM semantic verification, and (3) negation-based disproof filtering strategies using off-the-shelf LLM-based provers. 
This approach reduces expert annotation costs by retaining 72.09\% of statements before manual verification while ensuring fidelity to the original natural-language problems.
Our evaluation of state-of-the-art LLM-based theorem provers reveals significant limitations: even the strongest models achieve only 16.46\% success rate under practical sampling budgets, exhibiting pronounced domain bias (\eg, excelling in algebra but failing in calculus) and over-reliance on simplified automation tactics.
Notably, we identify a counterintuitive inverse relationship between natural-language solution guidance and proof success in chain-of-thought reasoning scenarios, suggesting that human-written informal reasoning introduces noise rather than clarity in the formal reasoning settings.
We believe that FormalMATH provides a robust benchmark for benchmarking formal mathematical reasoning.

\end{abstract}

\vspace{-3mm}
\renewcommand{\arraystretch}{1.15}
\begin{center}
    {\fontfamily{pcr}\selectfont
        \begin{tabular}{rll}
            \projectpage & \textbf{Project Page} & \href{https://sphere-ai-lab.github.io/FormalMATH/}{[web]} \\[0.2mm]

            \github & \textbf{Github Repository} & \href{https://github.com/Sphere-AI-Lab/FormalMATH-Bench}{[code]} \\[0.2mm]
    
            \dataset & \textbf{Huggingface Dataset} & \href{https://huggingface.co/SphereLab}{[data]} \\
    
        \end{tabular}
    }
\end{center}
\vspace{-4mm}

\vspace{.5mm}
\section{Introduction}
\vspace{-1.5mm}
\begin{figure}[t]
  \centering\includegraphics[width=\textwidth]{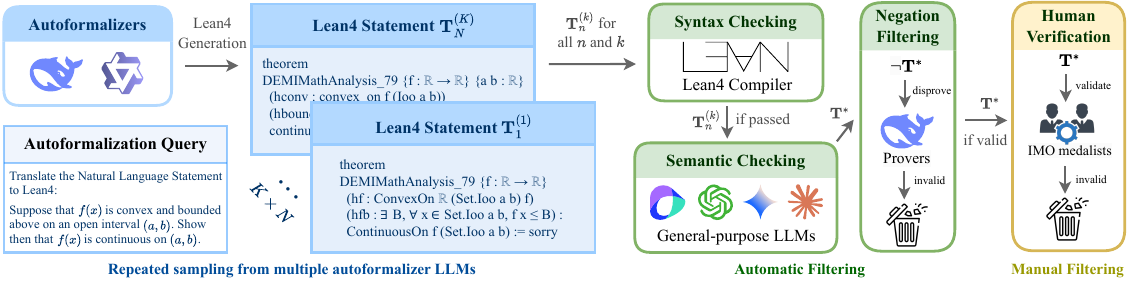}
  \vspace{-5.5mm}
  \caption{\small A human-in-the-loop pipeline for formal mathematical statement creation and filtering.}
  \label{fig:pipline}
  \vspace{-2mm}
\end{figure}
\vspace{2mm}

Formal mathematical reasoning (FMR)~\cite{yang2024formal} represents a specialized form of mathematical practice grounded in formal systems~\cite{leino2010dafny, mathlib, barras1997coq}, which provides a rigorous axiomatic framework essential for automated proof validation.
However, FMR is inherently challenging for humans.
For instance, the Liquid Tensor Experiment~\cite{scholze2022liquid} and the Polynomial Freiman-Ruzsa Conjecture~\cite{pfr_github} have taken years of effort by human experts to formalize and yet remain incomplete.
Recent works have leveraged self-supervised learning ~\cite{polu2020generative}, 
chain-of-thought (CoT) finetuning ~\cite{xin2024deepseek}, and scalable tree-search~\cite{xin2025bfs} to explore complex proof strategies, demonstrating the significant potential of large language models (LLMs) for FMR.
While there are several formal mathematics benchmarks, such as MiniF2F~\cite{zheng2021minif2f} and ProofNet~\cite{azerbayev2023proofnet} that are widely used to evaluate the FMR capabilities of LLMs, they still present a few critical limitations: (1) Scope limitation: Existing benchmarks are narrowly scoped. 
For instance, MiniF2F is restricted to high school-level algebra and number theory, while ProofNet focuses narrowly on undergraduate-level analysis and algebra. 
Their narrow scopes limit the capacity to evaluate holistic FMR capabilities across diverse mathematical domains.
(2) Dataset size: Formal mathematics benchmarks remain relatively small in scale. 
MiniF2F contains merely 244 problems in its test set, and ProofNet includes only 186.
This constrains benchmarking robustness and hinders the development of generalizable FMR systems. (3) Performance Saturation: State-of-the-art theorem provers, such as Kimina-Prover~\cite{wang2025kiminaproverpreviewlargeformal}, now achieve success rates exceeding 80.7\%, signaling that existing benchmarks may be nearing their practical utility limits.

\vspace{1.5mm}

To address these limitations, we introduce FormalMATH --- a large-scale Lean4~\cite{moura2021lean}-based benchmark containing 5,560 formally verified mathematical statements. 
FormalMATH includes a broad spectrum of mathematical domains, such as algebra, geometry, calculus, number theory, discrete mathematics, and more, while simultaneously spanning multiple difficulty levels, ranging from high school olympiad problems to undergraduate-level theorems (see Figure~\ref{fig:domain} for an overview).
The development of FormalMATH presents two primary challenges: (1) Autoformalization difficulty: limited concurrent tools open-sourced for robustly translating natural-language problems into precise Lean4 statements, especially for advanced mathematical domains requiring strict semantic preservation, (2) Validating formal statements requires ensuring syntactic correctness (via Lean4 compiler checks) and semantic alignment with original problems—a dual requirement that remains technically demanding and time-intensive even for human experts.

\vspace{1.5mm}

Motivated by these challenges, we propose a human-in-the-loop framework (Figure~\ref{fig:pipline}) for constructing the FormalMATH benchmark. 
Our framework substantially reduces the manual annotation effort required to generate formal mathematical statements by integrating:
(1) Ensemble-based autoformalization: multi-LLMs for autoformalization via best-of-N sampling~\cite{wang2022self} and (2) Automated validation: A three-tiered pipeline ensures correctness --- compiler syntax validation \cite{repl}, Multi-LLMs semantic verification, and negation-based disproof to filter unprovable theorems.
This strategy minimizes human verification while achieving high accuracy, preserving 72.09\% of translated statements in FormalMATH.

\vspace{1.5mm}

We evaluate state-of-the-art LLM-based theorem provers on the FormalMATH benchmark, revealing significant challenges for these systems. 
For instance, the best-performing model --- Kimina-Prover~\cite{wang2025kiminaproverpreviewlargeformal} achieves only 16.46\% on the FormalMATH-Full dataset under the pass@32 metric, while BFS-Prover~\cite{xin2025bfs} attains just 11.13\% using a best-first search with a sampling budget of $1 \times 32 \times 100$.
Our analysis of these results yields several intriguing insights.
First, existing provers exhibit a pronounced domain bias, excelling primarily in high-school-level algebra and applied mathematics while struggling with other mathematical domains. 
This highlights critical gaps in their cross-domain generalizability. 
Second, the provers frequently reduce multi-step reasoning to single-tactic invocations (\eg, ``aesop''~\cite{limperg2023aesop} and ``linearith''), bypassing necessary deductive rigor.
Third, while CoT reasoning~\cite{wei2022chain} enhances performance on FormalMATH statements, adding natural language solutions reduces success rates, suggesting such guidance introduces ambiguity rather than clarity.
Our contributions include:

\vspace{1.5mm}

\begin{itemize}[leftmargin=*,nosep]
\setlength\itemsep{0.4em}
\item \textbf{A Large and Comprehensive Lean4 Benchmark}: We present FormalMATH, a benchmark of 5,560 formally verified mathematical statements spanning diverse subdomains. This dataset is 22.8$\times$ larger than the widely used MiniF2F benchmark.
\item \textbf{An Effective Human-in-the-Loop Autoformalization Pipeline}: We introduce a framework (Figure~\ref{fig:pipline}) integrating multi-LLM autoformalization, multi-LLM semantic verification, and negation-based disproof strategies to automate formalization while retaining 72.09\% accuracy before manual review. This reduces reliance on costly expert annotation and enables scalable Lean4 dataset construction.
\item \textbf{Comprehensive Evaluation of LLM-based Theorem Provers}: 
Our systematic evaluation reveals fundamental limitations in state-of-the-art theorem provers: (1) even the best-performing model achieve only a 16.46\% success rate on FormalMATH, (2) existing provers exhibit severe domain bias, performing well in areas like algebra but poorly in others such as calculus, (3) counterintuitive inverse relationship where providing natural language solution guidance decreased proof success rates in CoT scenarios.
These limitations suggest important potential directions for improving LLM-based provers.
\end{itemize}

\begin{figure}[t]
  \centering
    \captionsetup[subfigure]{belowskip=-5pt, aboveskip=2.5pt}
  \captionsetup[subfigure]{margin=25pt}
  \vspace{-2mm}
  \begin{subfigure}[b]{0.645\textwidth}
    \centering
    \includegraphics[width=.99\linewidth]{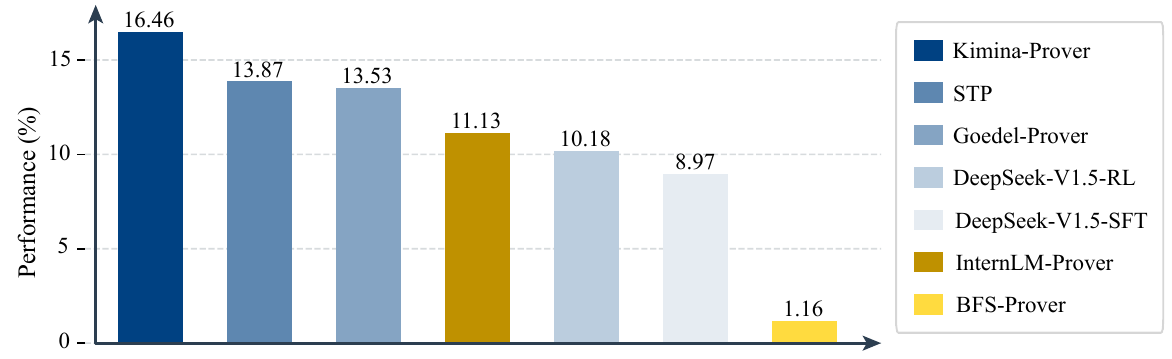}
    \caption{\small Performance of current provers on FormalMATH}
    \label{fig:main:sota}
  \end{subfigure}
  \hfill 
  \begin{subfigure}[b]{0.345\textwidth}
    \centering
    \includegraphics[width=1\textwidth]{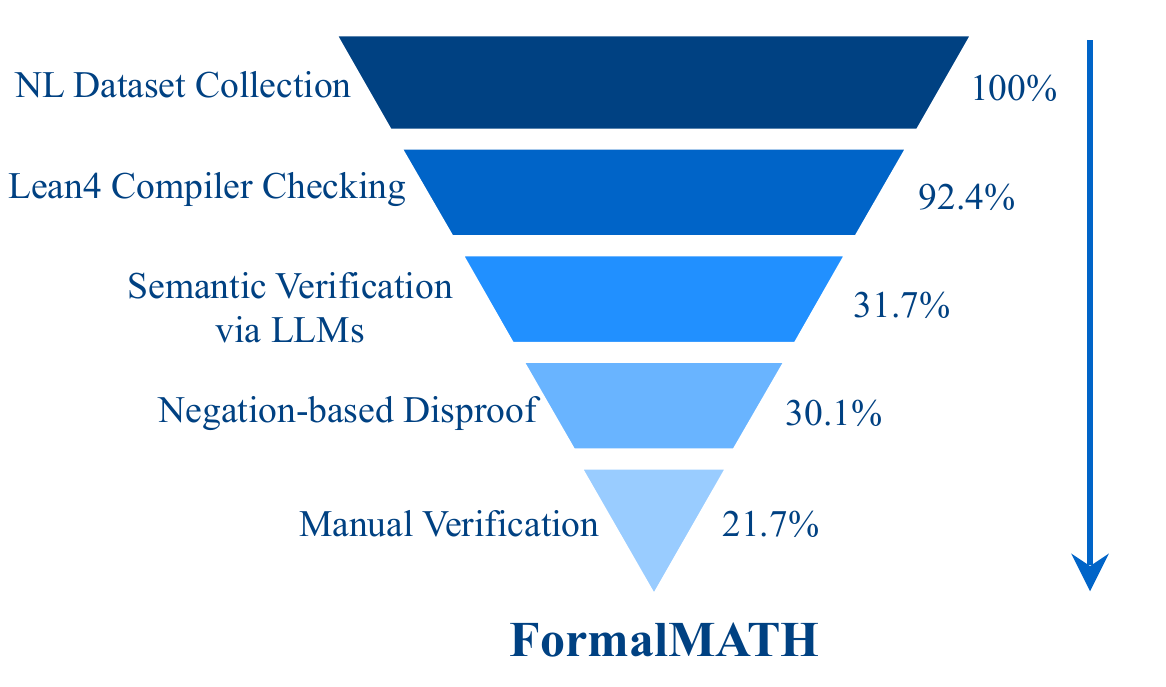}
    \caption{\small Data preservation rate}\label{fig:main:filter}
  \end{subfigure}
  \caption{\small (a) Performance comparison of existing theorem provers on the full FormalMATH benchmark. Results show Pass@1$\times$32$\times$100 accuracy for best-first-search-based (BFS) methods, including BFS-Prover and InternLM-Prover, and Pass@32 accuracy via single-pass generations (SPG) for the other provers, including Kinima-Prover, STP, Goedel-Prover, DeepSeek-V1.5-RL and DeepSeek-V1.5-SFT.
  (b) Funnel chart illustrating the percentage of data that is preserved after each filtering stage in our human-in-the-loop autoformalization pipeline.}
  \label{fig:main}
  \vspace{-1mm}
\end{figure}

\section{Related Work}
\textbf{Autoformalization} refers to the task of automatically translating informal mathematics (\eg, problem statements from sources like \cite{cobbe2021training,Yu2024MetaMath}) into formal mathematics (\eg, Lean4~\cite{moura2021lean} or Isabelle~\cite{nipkow2002isabelle}).
Recent work has leveraged LLMs~\cite{OpenAI} using two main paradigms: (1) In-context learning, where models generalize from examples provided within prompts~\cite{wu2022autoformalization, liu2023fimo, lu2024process}, and (2) Data-driven supervised finetuning, which uses carefully curated pairs (\eg, potentially augmented CoT~\cite{wei2022chain} via general-purpose LLMs) of natural and formal language to train autoformalization models~\cite{lin2025goedel, xin2024deepseek, wang2025kiminaproverpreviewlargeformal}.
A key challenge is how to validate the results of these autoformalization models. 
Previous evaluation metrics include machine translation metrics (\ie, BLEU~\cite{papineni2002bleu}) as employed in~\cite{wu2022autoformalization}, or process-guided annotation~\cite{lu2024process}.
Both approaches depend critically on comparing the LLM's output against a known ground-truth formalization.
Major Lean4 benchmarks, such as MiniF2F~\cite{zheng2021minif2f}, ProofNet~\cite{azerbayev2023proofnet}, and PutnamBench~\cite{tsoukalas2024putnambench}, rely entirely on manual formalization by human experts to formalize the mathematical statement. 
This costly process highlights significant scalability limitations.
FormalMATH addresses these constraints by introducing a simple yet effective human-in-the-loop approach, where a carefully designed multi-LLM automated filtering strategy precedes manual review, making the generation of formalized statements more efficient and scalable.

\vspace{1.5mm}

\noindent\textbf{Formal Mathematical Reasoning.}
Current LLM-based Formal Mathematical Reasoning (FMR) methods~\cite{yang2024formal} differ substantially in their computational frameworks.
The predominant approach in FMR is proof search~\cite{polu2020generative, alphaproof2024ai, yang2023leandojo}, which generates proofs by combining tactic generation with search algorithms across evolving proof states.
Representative search strategies include best-first search~\cite{xin2025bfs, polu2020generative, wu2024internlm2}, Monte-Carlo tree search~\cite{coulom2006efficient, kocsis2006bandit, xin2024deepseek}, and Lean-STAR sampling~\cite{lin2024lean}.
While this approach ensures invalid tactics are immediately rejected through compiler verification, it inherently constrains the model's capacity for strategic reasoning and requires substantial computational resources to validate intermediate proof steps.
Alternatively, single-pass generation (SPG) methods (\eg, \cite{xin2024deepseek, lin2025goedel, dong2025beyond}) utilize LLMs to generate entire proofs directly. 
These methods then typically employ techniques like best-of-N sampling to scale up test-time computation, often achieving results comparable to proof-search methods.
As a SPG method, Kimina-prover~\cite{wang2025kiminaproverpreviewlargeformal} employs long-CoT~\cite{guo2025deepseek} with a think prompt template during reinforcement learning~\cite{team2025kimi}, achieving impressive performance.
Section~\ref{exp:formalall} compares various proof search and SPG methods on FormalMATH.

\vspace{2mm}
\setlength{\columnsep}{11pt}
\begin{wraptable}{r}{0.48\textwidth}
\setlength{\abovecaptionskip}{3pt}
\setlength{\belowcaptionskip}{3pt}
\setlength{\tabcolsep}{2pt}
\renewcommand{\arraystretch}{1.26}
\scriptsize
\centering
\begin{tabular}{lcc}
\specialrule{0em}{0pt}{-7pt}
\textbf{Benchmark} & \textbf{\# Problems} &\textbf{Difficulty} \\
\shline
MiniF2F~\cite{zheng2021minif2f}& 244 & Olympiad \\
ProofNet~\cite{azerbayev2023proofnet} & 186 & Undergraduate (UG) \\
FIMO~\cite{liu2023fimo} & 149 & Olympiad \\
PutnamBench~\cite{tsoukalas2024putnambench}& 522 & Olympiad \\
ProverBench~\cite{ren2025deepseek} & 325 & Olympiad \\
\rowcolor{Gray} FormalMATH & \textbf{5,560} & \textbf{Olympiad \& UG} \\
\specialrule{0em}{0pt}{0pt}
\end{tabular}
\caption{\small Comparison of existing Lean4 benchmarks.}
\label{tab:difficulty_comparison}
\vspace{-1mm}
\end{wraptable}

\noindent\textbf{Formal Theorem Proving Benchmarks.} 
Benchmarks for assessing Lean4-based theorem-proving capabilities can be categorized based on whether they use off-the-shelf formal proofs.
Benchmarks derived from existing libraries, such as LeanDojo~\cite{yang2023leandojo}, extract proofs and theorems from the off-the-shelf Lean Mathlib library~\cite{mathlib}.
In contrast, benchmarks without pre-formalized proofs operate under a different paradigm. 
Instead of providing reference proofs, these benchmarks present only formalized problem statements, often derived from informal mathematics. Proving systems are used to generate a proof from scratch, the validity of which is then verified using the Lean compiler~\cite{repl}.
As shown in Table~\ref{tab:difficulty_comparison}, representative benchmarks include:
(1) MiniF2F~\cite{zheng2021minif2f}, which compiles 244 competition-level problems from AMC, AIME, and IMO in its test dataset,
(2) ProofNet~\cite{azerbayev2023proofnet}, which comprises 186 problems from undergraduate-level analysis and algebra,
(3) FIMO~\cite{liu2023fimo}, which contains 149 IMO shortlist problems, and 
(4) PutnamBench~\cite{tsoukalas2024putnambench}, which is a benchmark of 522 Lean4 problems from the Putnam competition.
FormalMATH also falls into this latter category (requiring new proof completion), comprising 5,560 diverse problems formalized from high-school competition-level sources (\eg, Omni-Math~\cite{gao2024omni} and BlueMO~\cite{bluemo2024}) and undergraduate-level problems (\eg, U-Math~\cite{chernyshev2024u}, Hardmath~\cite{fan2024hardmath}, and DEMIMATH~\cite{demidovich1964problems}).

\section{\underline{FormalMATH}: A Large Formal Mathematical Reasoning Benchmark}

\subsection{Overall Dataset Statistics}
\label{exp:dataset}
\begin{figure}[t!]
  \centering
  \setlength{\abovecaptionskip}{4pt}
  \setlength{\belowcaptionskip}{4pt}
  \vspace{-1mm}
  \includegraphics[width=0.99\textwidth]{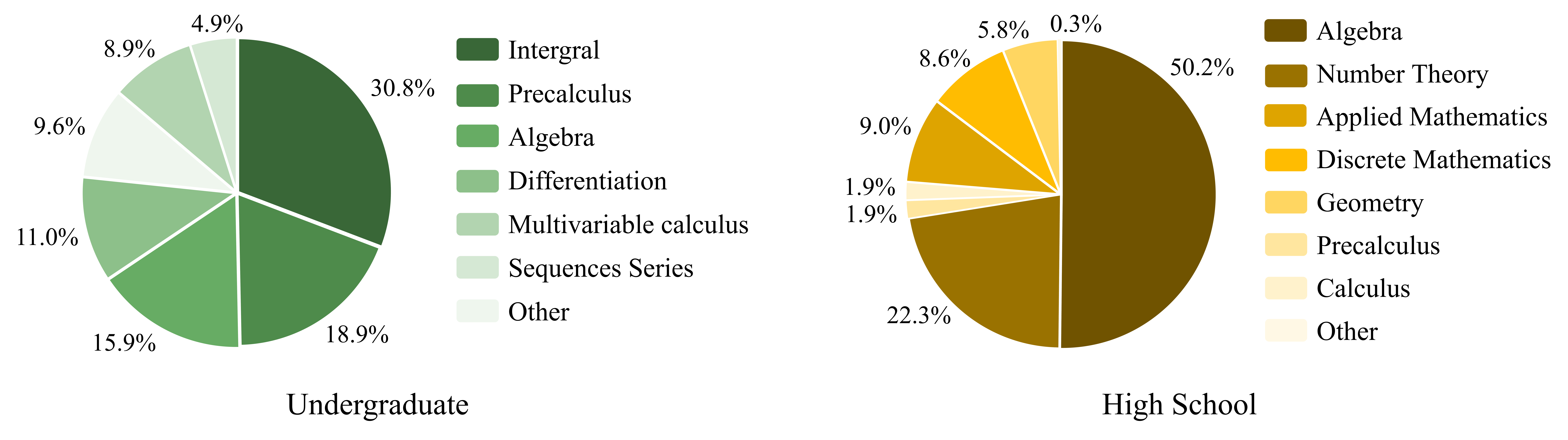}
  \caption{\small The distribution of mathematical domains in the full set of FormalMATH.}
  \label{fig:domain}
  \vspace{-2.5mm}
\end{figure}

FormalMATH is a rigorously validated Lean4 benchmark comprising 5,560 mathematical statements, each independently verified through a hybrid pipeline of multi‑LLM semantic verification and careful review by Olympiad‑level experts. Figure~\ref{fig:pipline} gives the overall autoformalization pipeline.
Figure~\ref{fig:main:filter} depicts the sequential validation process and the preservation rates at each stage. 
We list all data sources that contribute to FormalMATH in Appendix~\ref{app:data_source}.
The problems span a broad difficulty spectrum, from high‑school competition questions in disciplines such as algebra, number theory, discrete mathematics, and geometry, to undergraduate challenges in specialized areas including calculus (integration and differentiation), linear and abstract algebra, sequences and series.
Figure~\ref{fig:domain} provides the distribution of topic domains.
Appendix~\ref{sec:example} gives examples of the formalized Lean4 statements in FormalMATH.

\subsection{The Proposed Human-in-the-loop Pipeline for Data Collection and Filtering}
\label{exp:human}
\noindent \textbf{Supervised Fine-tuning.}
During the development of FormalMATH, we find that mature, open‑source autoformalization tools are scarce. To fill this gap, we build our own pipeline on top of two types of LLMs: coding‑specialized LLMs (\eg, Qwen2.5‑7B‑Coder~\cite{bai2023qwen}) and pre‑trained theorem‑proving LLMs (\eg, Deepseek‑prover‑base~\cite{xin2024deepseek}). We then generate training data by having a general‑purpose LLM (\eg, GPT‑4~\cite{OpenAI}) iteratively translate natural‑language statements into Lean4 statements. Each candidate statement is then passed to the Lean4 compiler, and only those that are type‑checked will be kept. This straightforward ``compile‑and‑filter'' strategy yields a high‑quality corpus of 9,260 paired training examples, which is eventually used to finetune our own autoformalization models.
\vspace{2mm}

\noindent \textbf{Autoformalization.} For each of the $K$ autoformalizers (implemented by LLMs), we employ a best-of-N sampling strategy~\cite{wang2022self} to generate $N$ formal candidate statements $\mathbf{T}_{n}^{(k)}$, where $k \in \{1, \dots, K\}$ denotes the autoformalizer index, and $n \in \{1, \dots, N\}$ represents the candidate statement index of the $k$-th autoformalizer. 
All candidate statements $\mathbf{T}_{n}^{(k)}$ are first validated for syntactic correctness using the Lean4 compiler. 
Only syntactically valid statements are preserved for subsequent semantic verification.

\vspace{2mm}

\noindent \textbf{Semantic Verification via LLMs.}
We implement a semantic verification strategy based on multiple powerful general-purpose LLMs (\eg, o1-mini~\cite{jaech2024openai}, claude-3.5-Sonnet) to evaluate semantic alignment between natural language mathematics problems and their Lean4 formalizations.
Each model employs chain-of-thought reasoning (See the prompt in Section~\ref{sec:semantic}) to complete the following procedures: (1) back-translate Lean4 statements into natural language, (2) compare reconstructed descriptions with original problems, and (3) provide binary judgments (\ie, aligned/misaligned). 
Importantly, only Lean4 statements that passed semantic verification performed by all the LLMs would be collected.
This strategy is guided by the insight that translating Lean4 statements to natural language is a much easier task than the reverse process, and general-purpose LLMs excel at understanding natural language phrasings~\cite{wu2022autoformalization}.
Overall, this procedure filters out 60.7\% of syntactically correct but semantically misaligned statements (\ie, from 92.4\% to 32.7\%).
Interestingly, we find distinct consensus patterns across problem difficulty levels -- around 30\% unanimous agreement rate for high school competition problems and significantly lower consensus for undergraduate-level formalizations (\eg, 4.63\% on HardMath).

\vspace{1mm}

\begin{tcolorbox}[
    title=\textbf{\scriptsize Example 3.2: Negation-Based Disproof Protocol to Filter out Non-provable Statements},
    colframe=c3,
    colback=c4!5,
    boxrule=0.5pt,
    fonttitle=\scriptsize, %
    fontupper=\scriptsize, %
    breakable,
]
\scriptsize
\vspace{0.2em}
\textbf{\scriptsize Original Lean4 Statement:}

\begin{lstlisting}[frame=single, basicstyle=\scriptsize]
import Mathlib

def refBase (n : ℕ) : Prop :=
  ∀ k l, 0 < k → 0 < l → k < n → l < n →
  (k | n → l | n → (2 * k - l | n ∨ 2 * l - k | n))

theorem olymid_ref_base_1120 : {n | 1 < n ∧ refBase n} = {6, 9, 15} := by
  sorry
\end{lstlisting}
\par
\vspace{0.1\baselineskip}
\textbf{\scriptsize Negation-based Disproof by Contradiction Construction:} 

\lstset{
    frame=single,
    basicstyle=\scriptsize,
    xleftmargin=0.1pt,
    framexleftmargin=0.1pt,
    keywordstyle=\bfseries,
    morecomment=[l]{--},   %
    commentstyle=\color{gray}, 
    escapeinside={@!}{@!},     %
     moredelim=**[is][\color{red}]{@}{@},
     language=lean,
}

\begin{lstlisting}[frame=single, basicstyle=\scriptsize]
theorem olymid_ref_base_1120_negative : {n | 1 < n ∧ refBase n} ≠ {6, 9, 15} := by
  simp (config := { decide := true }) [refBase]
  simp only [Set.ext_iff, Set.mem_setOf_eq, Set.mem_insert_iff, Set.mem_singleton_iff]
  intro h
  have h1 := h 7
  simp (config := { decide := true }) at h1
  obtain ⟨k, hk0, l, hl0, hk, hl, hkd, hld, h1, h2⟩ := h1
  interval_cases k <;> interval_cases l <;> simp_all (config := {decide := true})
\end{lstlisting}
\label{textbox:negprove}
\end{tcolorbox}
\vspace{1mm}

\noindent \textbf{Disproving a Statement by Proving Its Negation.}
Inspired by the Law of the Excluded Middle (LEM~\cite{wiki:excluded_middle}), we further filter out certain non‑provable formalizations using off‑the‑shelf LLM‑based provers (\eg, DeepSeek‑Prover‑V1.5).
For any formalized statement $\mathbf{T}_{n}^{(k)}$, we perform the following steps:
(1) construct logical negation: construct its logical negation by applying transformation rules such as De Morgan dualization~\cite{wiki:demorgan_laws} to generate $\neg \mathbf{T}_{n}^{(k)}$, and 
(2) automated proof attempts: perform automated proof attempts on $\neg \mathbf{T}_{n}^{(k)}$ within the formal system $\mathcal{S}$ (\ie, Lean4 compiler). 
A successful proof of $\neg \mathbf{T}_{n}^{(k)}$ implies that the original statement $\mathbf{T}_{n}^{(k)}$ cannot hold on $\mathcal{S}$. 
Example \ref{textbox:negprove} illustrates the Lean 4 formalization of a number‑theoretic conjecture and its negation.
By constructing the negation of a statement and applying an LLM-based prover for disproof, the system identifies inconsistencies through boundary case testing (\eg, n = 7) and derives contradictions via systematic case analysis (\ie, \texttt{interval\_cases}). This strategy has filtered out a few unprovable statements, accounting for 1.6\% of the total statements.

\vspace{1.5mm}

\setlength{\intextsep}{0pt}
\setlength{\columnsep}{11pt}
\begin{wraptable}{r}{0.26\textwidth} %
\setlength{\abovecaptionskip}{3pt}
\setlength{\belowcaptionskip}{6pt}
\setlength{\tabcolsep}{12pt} %
\renewcommand{\arraystretch}{1.26}
\scriptsize
\centering
\begin{tabular}{@{}lc@{}}
\textbf{Item} & \textbf{Value} \\
\shline
\# Annotators & 12 \\
Preservation rate & 72.09\% \\
Cost/statement & \$6.89 \\
Total duration & 22 days \\
\end{tabular}
\caption{\small Annotation statistics.}
\label{tab:verif_metrics}
\end{wraptable}

\noindent \textbf{Expert Verification.}
We have recruited 12 International Mathematical Olympiad medalist-level human experts to manually check the semantic alignment fidelity between natural language statements and their Lean4 formalizations. Table~\ref{tab:verif_metrics} shows some relevant information about the human validation stage.
Our results show that our multi-LLM autoformalization and validation pipeline delivers substantial efficacy, retaining 72.1\% of statements from the last stage of LLM-based semantic verification (from 30.1\% to 21.7\%) while significantly reducing manual verification efforts.
In total, we have successfully formalized 21.7\% of syntactically and semantically correct mathematical statements from a diverse collection of mathematical problems collected from multiple data sources. See Appendix~\ref{app:data_source},\ref{app:errortype} for more details.

\section{Experiments and Discussions}

\subsection{Evaluating Formal Theorem Provers on FormalMATH}
\label{exp:formalall}

\begin{table}[!t]
\setlength{\abovecaptionskip}{4pt}
\setlength{\belowcaptionskip}{0pt}
\setlength{\tabcolsep}{18pt}
\renewcommand{\arraystretch}{1.1}
\small
\centering
\vspace{-2mm}
    \begin{tabular}{lcc}
        \textbf{Method}  & \textbf{Sampling budget} & \textbf{Pass@$K$ (\%)} \\
        \shline
        \multicolumn{3}{c}{\textit{Best-First Tree Search Methods}} \\[0.5mm]
        \multirow{5}{*}{BFS(DeepSeek-Prover-V1.5-RL) \cite{xin2024deepseek}}
        &$1\times32\times100$&  $4.91$ \\
        &$4\times32\times100$&  $10.29$ \\
        &$8\times32\times100$&  $12.16$ \\
        &$16\times32\times100$&  $14.96$ \\
        &$32\times32\times100$&  $17.41$ \\
        \hline
        \multirow{5}{*}{BFS(InternLM-V2.5) \cite{wu2024internlm2}}
        &$1\times32\times100$&  $7.87$ \\
        &$4\times32\times100$&  $15.79$ \\
        &$8\times32\times100$&  $20.02$ \\
        &$16\times32\times100$&  $22.74$ \\
        &$32\times32\times100$&  $25.65$ \\
        \hline
        \multirow{5}{*}{BFS(BFS-Prover) \cite{xin2025bfs}}
        &$1\times32\times100$&  $27.10$ \\
        &$4\times32\times100$&  $34.04$ \\
        &$8\times32\times100$&  $37.56$ \\
        &$16\times32\times100$&  $41.75$ \\
        &$32\times32\times100$&  $45.88$ \\
        \shline
        \multicolumn{3}{c}{\textit{Single-Pass Generation Methods}} \\[0.5mm]
        Kimina-Prover-7B~\cite{wang2025kiminaproverpreviewlargeformal} & 32 & $48.94$ \\
        \hline
        \multirow{6}{*}{STP~\cite{dong2025beyond}}
        &32&  $48.59$ \\
        &128&  $50.35$ \\
        &512&  $51.45$ \\
        &1024&  $52.03$ \\
        &2048&  $52.60$ \\
        &3200&  $53.17$ \\
        \hline
        \multirow{6}{*}{DeepSeek-Prover-V1.5-SFT~\cite{xin2024deepseek}}
        &$32$&  $40.40$ \\
        &128&  $42.11$ \\
        &512&  $44.17$ \\
        &1024&  $45.08$ \\
        &2048&  $46.12$ \\
        &3200&  $46.82$ \\
        \hline
        \multirow{6}{*}{DeepSeek-Prover-V1.5-RL~\cite{xin2024deepseek}}
        &32&  $47.98$ \\
        &128&  $48.75$ \\
        &512&  $49.27$ \\
        &1024&  $49.68$ \\
        &2048&  $50.08$ \\
        &3200&  $50.35$ \\
        \hline
        \multirow{6}{*}{Goedel-Prover~\cite{lin2025goedel}}
        &32&  $46.70$ \\
        &128&  $48.02$ \\
        &512&  $48.68$ \\
        &1024&  $49.04$ \\
        &2048&  $49.20$ \\
        &3200&  $49.41$ \\
        \hline
        \multirow{1}{*}{Ensemble of All SPG Methods} &$4\times3200$&54.11 \\

    \end{tabular}    
    \caption{\small Performance comparison of theorem prover LLMs on FormalMATH-Lite.}
    \label{tab:main-results}
    \vspace{-1mm}
\end{table}

\noindent \textbf{LLM-based Prover Settings.}
We focus on the following two different proof-generation approaches: 

\vspace{2mm}

\begin{itemize}[leftmargin=*,nosep]
\setlength\itemsep{0.4em}
\item \textbf{Best-First Tree-Search (BFS) Methods.} Each node in the search tree represents an intermediate proof state, and a heuristic scoring function assigns a priority to each node. We evaluate three baseline models under this category: BFS-Prover \cite{xin2025bfs}, DeepSeek-Prover-V1.5-RL \cite{xin2024deepseek}, and InternLM-V2.5-Prover \cite{wu2024internlm2}.
\item \textbf{Single-Pass Generation Methods.} The models under this category generate a complete proof in one pass, without iterative refinement or explicit intermediate states. In our paper, we consider the following baseline models: STP \cite{dong2025beyond}, DeepSeek-Prover-V1.5-SFT~\cite{xin2024deepseek}, DeepSeek-Prover-V1.5-RL~\cite{xin2024deepseek}, Goedel-Prover \cite{lin2025goedel}, and Kimina-Prover-7B \cite{wang2025kiminaproverpreviewlargeformal}.
\end{itemize}

\vspace{2mm}

\noindent \textbf{Metrics.} 
We evaluate theorem provers using the Pass@$K$ metric, which measures the fraction of problems for which at least one valid proof is found among the top $K$ generated attempts.
(1) For BFS, $K = N \times S \times T$, where $N$ denotes the number of best-first search attempts, $S$ is the number of tactics proposed during each expansion, and $T$ is the total number of expansion iterations.
(2) For SPG, $K$ corresponds to the total number of complete proof trajectories sampled from the model.

\vspace{1.5mm}

\noindent \textbf{Prompts.} 
In the experiments, we only consider vanilla generation strategies (see Example~\ref{textbox:prompt_vanilla}), where models directly generate Lean4 proof without explicit requirement of chain-of-thought (CoT) rationales (natural language thoughts interleaved with Lean4) or augmenting with natural language solutions.

\vspace{1.5mm}

\begin{wrapfigure}{r}{0.5\textwidth}
  \centering
  \setlength{\abovecaptionskip}{4pt}
  \setlength{\belowcaptionskip}{5pt}
  \includegraphics[width=0.5\textwidth]{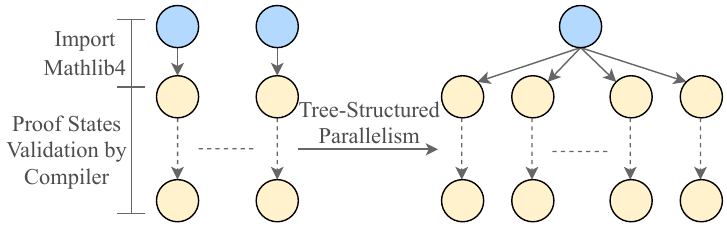}
  \caption{\small Our efficient Lean4 verifier implementation.}
  \label{fig:tree_verifier}
\end{wrapfigure}

\noindent \textbf{Verifier.} 
In Lean4, the correctness of proofs is verified by the compiler~\cite{repl}. 
However, verifying individual proofs is often time-consuming, largely due to the significant overhead associated with importing the Mathlib4 library~\cite{mathlib}. 
To mitigate this inefficiency, we use a tree-structured parallelism approach (see Figure~\ref{fig:tree_verifier}).
In this implementation, a parent thread manages the root node, which handles the computationally intensive import operations of Mathlib4. 
Concurrently, child threads process subsequent nodes in parallel, each corresponding to an individual proof. 
By centralizing the costly import operation at the root, redundant overhead is eliminated, and resources are efficiently allocated to parallelize proof verification.
This simple trick effectively optimizes test-time efficiency by avoiding repeated computational overhead, ensuring scalable and efficient utilization of computational resources.

\begin{figure}[!h]
  \centering
  \setlength{\abovecaptionskip}{4pt}
\setlength{\belowcaptionskip}{2pt}
\vspace{5mm}
  \includegraphics[width=1\textwidth]{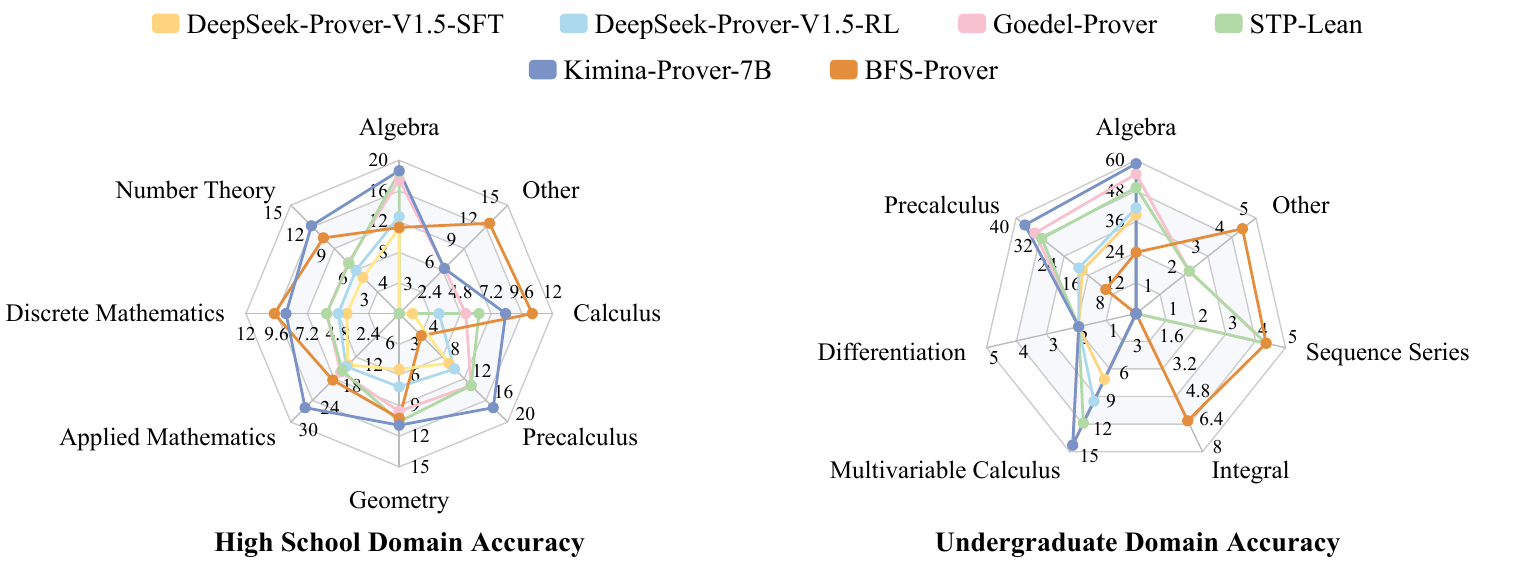}
  \caption{\small Breakdown of accuracy by mathematical domain within FormalMATH.} 
  \label{fig:category_result}
\end{figure}

\vspace{4mm}

\noindent \textbf{Finding 1: Existing LLM-based Provers Are Still Far from Solving FormalMATH.} 
Current LLM-based theorem provers demonstrate unsatisfactory performance on the FormalMATH benchmark under modest sampling budgets.
Specifically, one of the current strongest SPG methods, Kimina-Prover, achieves a mere 16.46\% under Pass@32, while the best BFS method, BFS-Prover, attains only 11.13\% Pass@1$\times$ 32 $\times $100, demonstrating the underlying difficulties of FormalMATH.
Notably, both methods use Qwen2.5-Math-7B as their base model but the performance differs dramatically: the former distills curated long-CoT proof traces from a larger LLM-based oracle, and the latter relies on expert iteration via BFS to iteratively enhance the LLM's Lean4 proving abilities.

Methods built upon DeepSeek-Prover-V1.5 exhibit a performance hierarchy that underscores the fundamental limitations of common post-training strategies nowadays. 
While the DeepSeek-V1.5-SFT baseline achieves 8.97\% accuracy, its reinforcement learning (RL) variant improves only marginally to 10.18\%--a mere +1.21\% gain that exposes the diminishing returns of rule-based sparse reward shaping in complex theorem spaces. 
However, another more sophisticated training paradigm, STP's self-play curriculum learning, achieves
13.87\% (+4.89\% over SFT) while Goedel-Prover's expert iteration reaches 13.53\% (+4.55\% over SFT).
Overall, these low success rates on FormalMATH underscore that current limitations of LLM-based provers: 
(1) reward sparseness: relying solely on binary rewards makes generalization to complex problems difficult, and techniques like intrinsic rewards may better guide exploration and skill acquisition.
(2) combinatorial search complexity: brute-force search and dependency on limited successful reasoning traces to RL and expert iteration affects sample efficiency and effective exploration.

\vspace{1.5mm}

\begin{figure}[tbp]
  \centering
  \vspace{-1mm}
  \setlength{\belowcaptionskip}{0pt}
  \captionsetup[subfigure]{belowskip=-4pt, aboveskip=3pt}
  \captionsetup[subfigure]{margin=25pt}
  \begin{subfigure}[b]{0.48\textwidth}
    \centering

    \includegraphics[width=\textwidth]{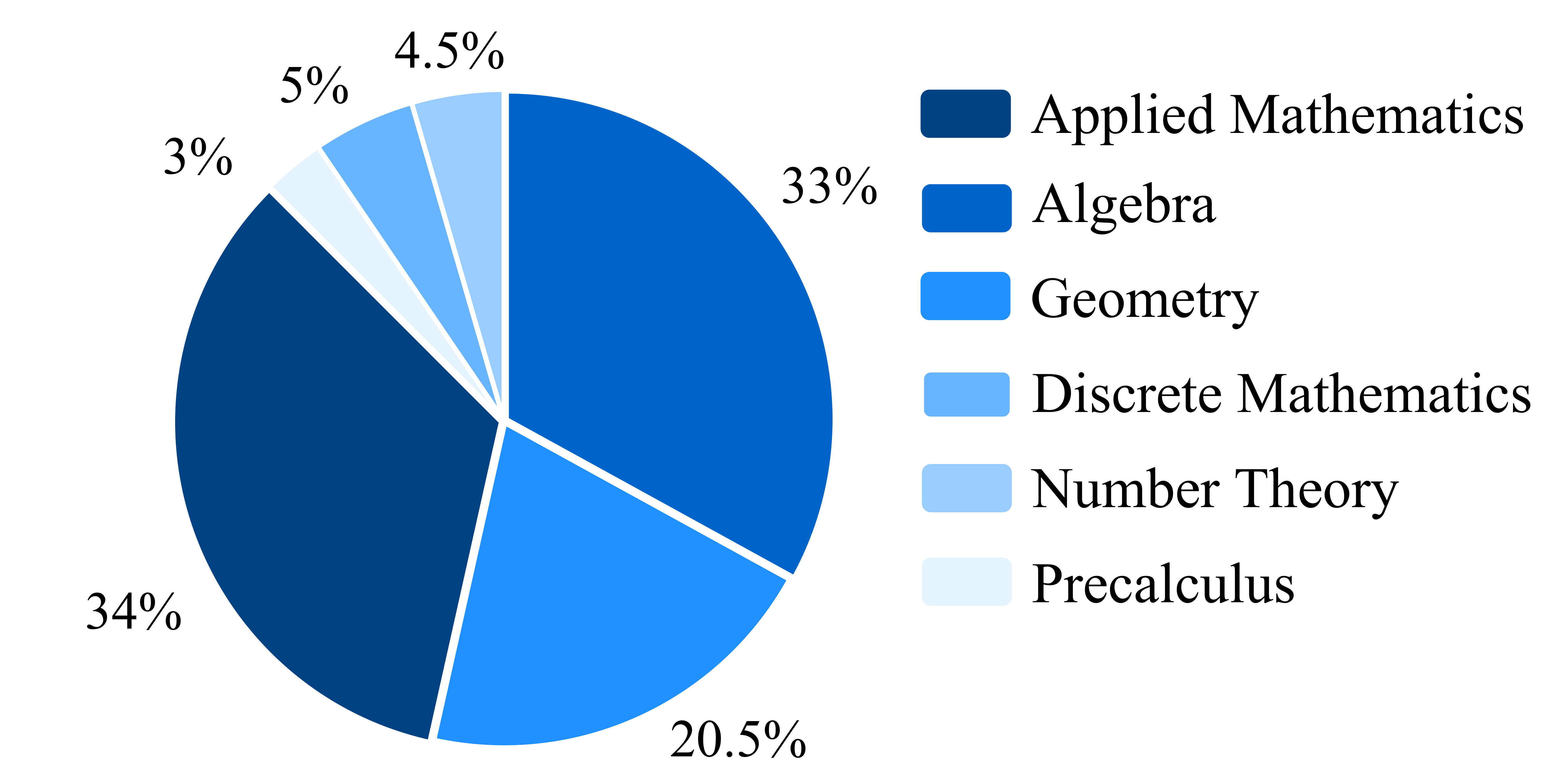}
    \caption{\small Training Domains of Goedel-Prover}
    \label{fig:godel_pie}
  \end{subfigure}
  \hfill 
  \begin{subfigure}[b]{0.49\textwidth}
    \centering
    \includegraphics[width=\textwidth]{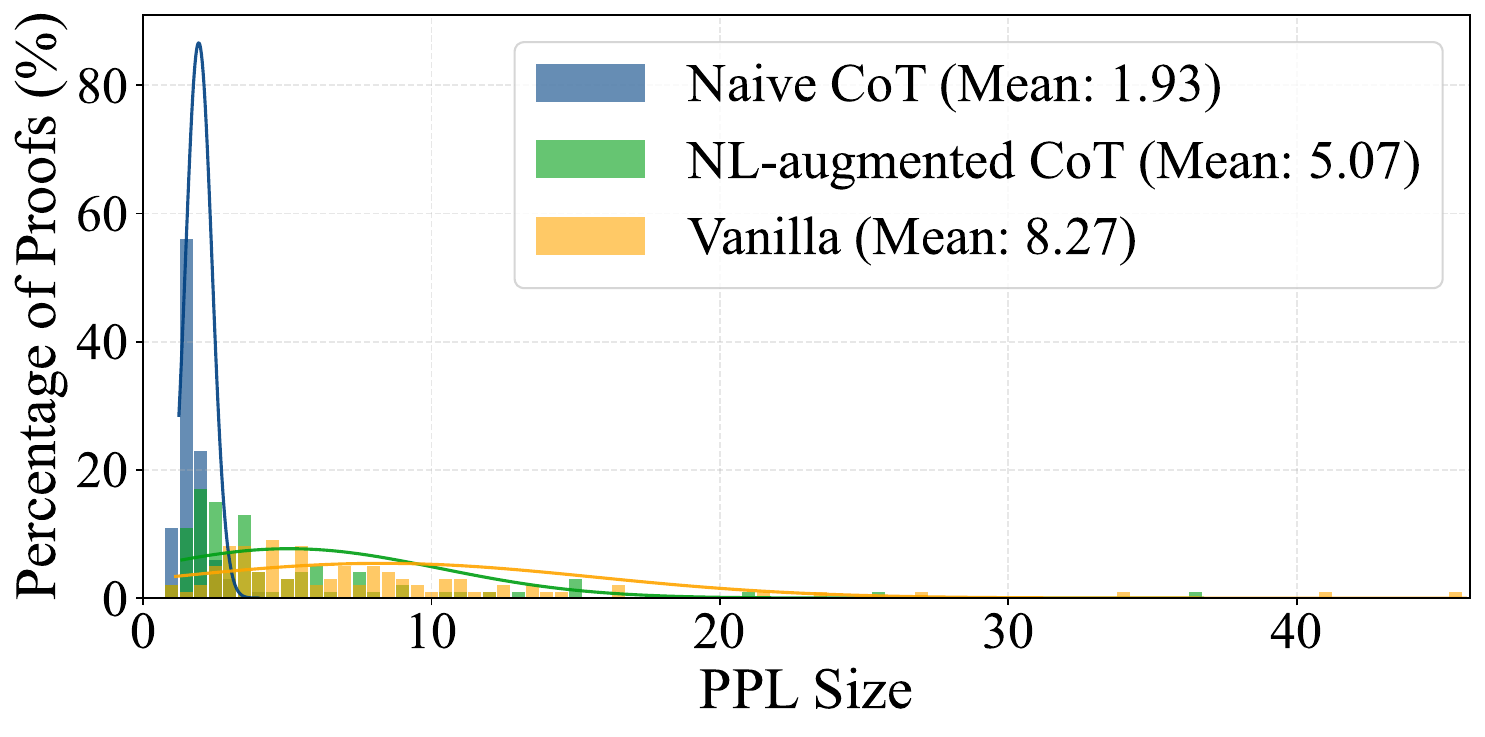}
    \caption{\small Perplexity of DeepSeek-V1.5-SFT}
    \label{ppl}
  \end{subfigure}
  \caption{\small (a) The mathematical domain distribution of Goedel-Prover's training dataset. (b) The perplexity distribution of Deepseek-V1.5-SFT across various proof generation modes.}
  \label{fig:ppl}
\end{figure}

\noindent \textbf{Finding 2: Provers' Unbalanced Performance Across Mathematical Domains of FormalMATH.}
Figure~\ref{fig:category_result} reveals significant domain bias in existing theorem provers. Under Pass@32, Godel-Prover achieves strong performance in algebra-related domains (\eg, 17.47\% in high school algebra and 50\% in undergraduate algebra) but performs poorly in calculus (5.21\%) and discrete mathematics (0\%). This imbalance persists at the undergraduate level, with success rates in precalculus (33.71\%) far exceeding those in differentiation (1.92\%) and integration (0\%).
We attribute this bias to the training data distributions. 
Using FormalMATH's domain categorization prompt (see Example~\ref{prompt:classify}), we analyzed Godel-Prover's training corpus by sampling 200 problems. As shown in Figure~\ref{fig:godel_pie}, the dataset disproportionately emphasizes applied mathematics and algebra (68\% combined), while discrete mathematics, number theory, and precalculus collectively constitute less than 5\%.

\subsection{Evaluating Test-time Scaling of Formal Theorem Provers on FormalMATH-Lite}
\label{exp:formallite}
Inspired by the recent success of test-time compute scaling~\cite{snell2024scaling,xiao2024verbalized,muennighoff2025s1,yu2025generating}, this section examines its impact on the formal mathematical reasoning capabilities of LLM-based theorem provers using our FormalMATH benchmark. To simplify, we only evaluate BFS and repeated sampling here.
To enable a systematic evaluation, we introduce FormalMATH-Lite, which is a curated subset of FormalMATH designed for efficient yet rigorous test-time scaling analysis. We compare state-of-the-art provers' performance on FormalMATH-Lite under varying sampling budgets, as shown in Table~\ref{tab:main-results}.

\vspace{1.5mm}
\noindent \textbf{FormalMATH-Lite.} 
Evaluating the full FormalMATH benchmark under large sampling budgets (\eg, Pass@3200) requires prohibitively high computational resources. To enable scalable yet rigorous analysis, we propose FormalMATH-Lite--a carefully selected subset of 425 problems (comprising 359 high school-level and 66 undergraduate-level problems) designed with two critical features:
(1) We utilize DeepSeek-V1.5-RL for outcome-driven difficulty assessment, evenly sampling solvable and unsolvable problems via constrained sampling budgets (\eg, Pass@32).
This balanced approach effectively highlights measurable scaling effects during test-time evaluation.
(2) Domain Distribution Alignment: This subset follows a mathematical domain distribution similar to the full FormalMATH benchmark (algebra, calculus, discrete mathematics, etc) using stratified sampling, ensuring sufficient coverage of core disciplines.
In Appendix~\ref{app:domain_lite}, we also provide the detailed distribution of FormalMATH-Lite.

\vspace{1.5mm}
\noindent \textbf{Experimental Settings.}
In this experiment, we maintain identical experimental configurations to Section~\ref{exp:formalall}--including models, prompts, etc, with one critical exception: sampling budget scales.
Section~\ref{exp:formalall} used constrained sampling budgets (\eg, Pass@32) due to computational resource limitations of the full FormalMATH benchmark.
Here, leveraging FormalMATH-Lite, we deploy expanded sampling budgets (\eg, up to Pass@3200 for SPG and Pass@32$\times$32$\times$100 for BFS).

\vspace{1.5mm}

\noindent \textbf{Finding 3: Subtle Performance Enhancement via Test-time Scaling.}
Table~\ref{tab:main-results} reveals limited returns when applying test-time scaling to formal theorem proving on FormalMATH.
For instance, STP achieves only a 4.58\% absolute improvement (from 48.59\% at Pass@32 to 53.17\% at Pass@3200) despite a 100 $\times$ sampling budget increase. 
While BFS-Prover demonstrates better scaling dynamics, attaining an 18.78\% gain (27.10\% via Pass@1$\times$32$\times$100 to 45.88\% via Pass@32$\times$32$\times$100), under a 32$\times$ budget expansion, however, it still underperforms relative to SGP methods.

Ensembling SPG methods (\ie, STP, Goedel-Prover, DeepSeek-V1.5-SFT, and DeepSeek-V1.5-RL) yields only marginal gains, from 53.17\% by STP alone to 54.11\% -- a mere 0.84\% uplift.
This is in sharp contrast to the near-linear scaling performance increments in informal reasoning~\cite{muennighoff2025s1}.
In informal mathematics, pseudo-continuous reward signals during sampling create pathways where imperfect reasoning chains, despite their logical flaws, can occasionally ``stumble'' into correct answers. This suggests that valid conclusions may emerge even when the intermediate steps aren't rigorously sound.

Formal theorem proving lacks such tolerance.
A single misplaced tactic or type error invalidates the entire proof trajectory, rendering incremental sampling ineffective. 
While verifier-guided proof search (\eg, BFS with access to intermediate proof states) theoretically mitigates this brittleness better than SPG methods, current implementations remain computationally impractical and lack scaling efficiency.

\subsection{CoT Can Enhance Model Capabilities on Formal Mathematical Reasoning}
\begin{figure}[tbp]
  \centering
  \vspace{-1mm}
  \setlength{\belowcaptionskip}{0pt}
  \captionsetup[subfigure]{belowskip=-6pt, aboveskip=2pt}
  \captionsetup[subfigure]{margin=25pt}
  \begin{subfigure}[b]{0.49\textwidth}
    \centering
    \includegraphics[width=\textwidth]{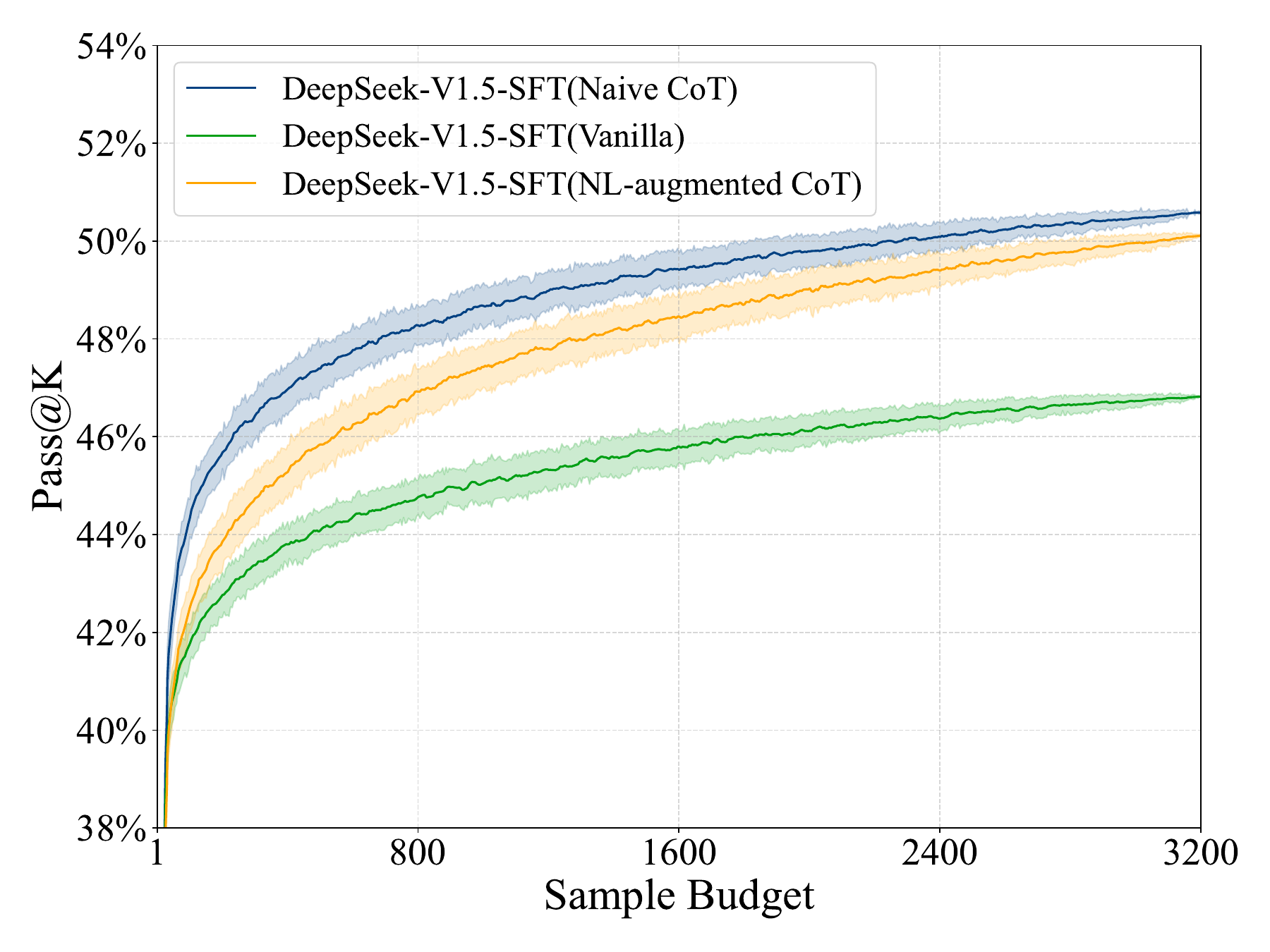}
    \caption{\small DeepSeek-V1.5-SFT}
    \label{a}
  \end{subfigure}
  \hfill 
  \begin{subfigure}[b]{0.49\textwidth}
    \centering
    \includegraphics[width=\textwidth]{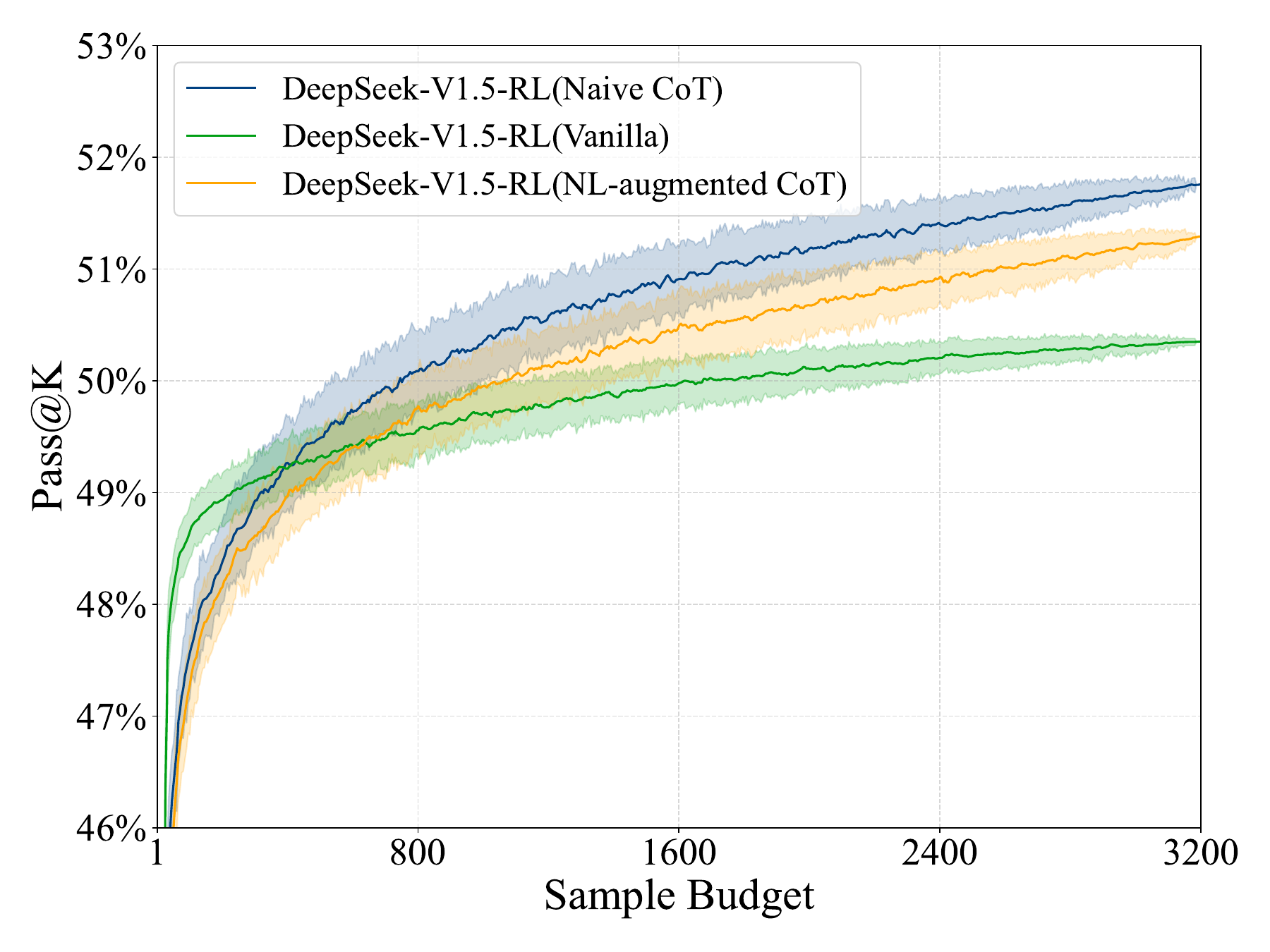}
    \caption{\small DeepSeek-V1.5-RL}
    \label{b}
  \end{subfigure}
  \caption{\small Pass@K accuracy curves for DeepSeek‑V1.5 provers across different reasoning configurations.}
  \label{fig:tts}
\end{figure}

In this section, we evaluate three different reasoning strategies in Lean4 proof generations: (1) naive CoT prompting (see Example~\ref{textbox:prompt_cot}), (2) NL-augmented CoT (see Example~\ref{textbox:prompt_nat}): CoT augmented with natural language solution example, and (3) vanilla generation strategies (see Example~\ref{textbox:prompt_vanilla}) via test-time scaling on FormalMATH-Lite (See Figure~\ref{fig:tts}).
Our goal is to \textit{measure whether---and to what extent---informal mathematical reasoning contributes to the rigor and effectiveness of subsequently derived formal proofs}.

\vspace{1.5mm}
\noindent \textbf{Experimental Setups.}
We evaluate DeepSeek‑Prover‑V1.5‑SFT and DeepSeek‑Prover‑V1.5‑RL (which are the only models explicitly trained with all three prompting strategies) on the FormalMATH‑Lite benchmark by applying test‑time scaling (up to Pass@3200).

\vspace{1.5mm}
\noindent \textbf{Finding 4: Naive CoT Outperforms Natural Language Guidance in Formal Theorem Proving.}
Across both SFT and RL configurations, we observe a consistent ranking of decoding strategies. Generally, naive CoT attains the highest Pass@K (from K equals 32 to 3200) accuracy, while NL-augmented CoT performs an intermediate position better than vanilla decoding.
For example, under $K=3200$, DeepSeek‑V1.5-SFT achieves 50.6\% with CoT and 49.2\% with NL‑augmented CoT and 47.0\% with vanilla decoding, and DeepSeek‑V1.5-RL achieves 51.7\%,  51.2\%, and 49.8\%, respectively.
On the other hand, it appears to be counterintuitive that NL-augmented CoT does not yield superior results compared to simple CoT. 
Figure~\ref{ppl} reveals a counterintuitive trend in perplexity distributions across prompting strategies: NL-augmented CoT consistently increases model uncertainty compared to naive CoT (\ie, mean perplexity from 1.93 to 5.07) across Lean4 problems.

\vspace{1mm}

In Example~\ref{textbox:dpsk_cot}, the failed NL‑augmented CoT proof reveals a fundamental error pattern: although the natural‑language outline and the Lean4 script target the same semantic goal, the high‑level sketch omits essential parameters and case distinctions that Lean’s tactics require. 
We hypothesize that this discrepancy stems from an intrinsic misalignment between the action space of informal, natural‑language reasoning and the tactic space of Lean4 formalization.

In this particular instance, the NL‑augmented CoT followed the NL solution by working on \texttt{modulo 7}, and asserting informally that $x^3 \; mod \;7 \in \{0,1,6\}$ and $y^4 \;mod\; 7\;\in \{0,1,2\}$ but does not materializes those assertions into the fifteen concrete \texttt{have $_{ \dots} =$ const} hypotheses branch that Lean4's decision procedures demand.
As a result, when the script invokes tactics (\ie, \texttt{omega}) reports the context simply lacks the linear congruences needed to derive a contradiction.

In contrast, naive CoT autonomously selects the larger prime modulus 13 without human-written prior, interleaves each residue‐case split with explicit tactic calls producing hypotheses like \texttt{have h : $x^3$ \% 13 = 5 := by simp [hxy\_mod]},
and then immediately discharges each branch with $\texttt{omega}$.
By reconstructing its own detailed, tactic‐level proof, CoT aligns semantic insight with low‑level proof obligations, guaranteeing that every subgoal carries the precise numeric constraints required for full automation--whereas the NL‑augmented approach, despite being semantically correct at a high level, leaves critical tactical steps unstated and thus fails to complete the proof.

\begin{tcolorbox}[
    title=\textbf{\scriptsize  Example 4.3: naive CoT outperforms NL-augmented CoT and vanilla strategies in formal theorem proving.},
    colframe=c3,
    colback=c4!5,
    boxrule=0.5pt,
    fonttitle=\scriptsize, %
    fontupper=\scriptsize, %
     breakable
]
\footnotesize
\vspace{0.2em}
\textbf{\scriptsize Naive CoT (Success):}
\begin{lstlisting}[frame=single, basicstyle=\scriptsize]
theorem algebra_68653_cot : ¬ ∃ x y : ℤ, x^3 + y^4 = 7 := by
  -- Assume there are integers x and y such that x^3 + y^4 = 7.
  intro h
  -- Extract the integers x and y from the assumption.
  rcases h with ⟨x, y, hxy⟩
  -- Consider the equation modulo 13.
  have hxy_mod : (x^3 + y^4) %
  -- Calculate the possible values of x^3 modulo 13.
  have x3_mod : x^3 %
    have : x %
    rcases this with (h | h | h | h | h | h | h | h | h | h | h | h | h) <;> simp [h, pow_succ, Int.mul_emod]
  -- Calculate the possible values of y^4 modulo 13.
  have y4_mod : y^4 %
    have : y %
    rcases this with (h | h | h | h | h | h | h | h | h | h | h | h | h) <;> simp [h, pow_succ, Int.mul_emod]
  -- Check that x^3 + y^4 cannot be congruent to 7 modulo 13.
  rcases x3_mod with (h1 | h1 | h1 | h1 | h1) <;> rcases y4_mod with (h2 | h2 | h2 | h2) <;> simp [h1, h2, Int.add_emod] at hxy_mod <;> omega

\end{lstlisting}
\par
\vspace{0.1\baselineskip}
\textbf{\scriptsize NL-agumented CoT (Not Success):} 

\lstset{
    frame=single,
    basicstyle=\scriptsize,
    xleftmargin=0.1pt,
    framexleftmargin=0.1pt,
    keywordstyle=\bfseries,
    morecomment=[l]{--},   %
    commentstyle=\color{gray}, 
    escapeinside={@!}{@!},     %
     moredelim=**[is][\color{red}]{@}{@},
     language=lean,
}

\begin{lstlisting}[frame=single, basicstyle=\scriptsize]
theorem algebra_68653_nat : ¬ ∃ x y : ℤ, x^3 + y^4 = 7 := by   /-
  We aim to show that there do not exist integers \( x \) and \( y \) such that \( x^3 + y^4 = 7 \). Suppose, for the sake of contradiction, that such integers exist. We would then have \( x^3 + y^4 = 7 \). We analyze the possible values of \( x^3 \mod 7 \) and \( y^4 \mod 7 \) to find a contradiction.
  1. **Lemma:** For any integer \( x \), \( x^3 \mod 7 \) can only be 0, 1, or -1.
     - Proof: By examining all possible values of \( x \mod 7 \), we find that \( x^3 \mod 7 \) is either 0, 1, or -1.
  2. **Lemma:** For any integer \( y \), \( y^4 \mod 7 \) can only be 0, 1, or 2.
     - Proof: By examining all possible values of \( y \mod 7 \), we find that \( y^4 \mod 7 \) is either 0, 1, or 2.
  3. **Combining the Lemmas:** If \( x^3 + y^4 = 7 \), then \( x^3 \mod 7 \) and \( y^4 \mod 7 \) must sum to 7 modulo 7. However, the possible values of \( x^3 \mod 7 \) are 0, 1, or -1, and the possible values of \( y^4 \mod 7 \) are 0, 1, or 2. None of these combinations can sum to 7 modulo 7, leading to a contradiction.
  Thus, we conclude that there do not exist integers \( x \) and \( y \) such that \( x^3 + y^4 = 7 \).
  -/
  -- Assume for contradiction that there exist integers x and y such that x^3 + y^4 = 7.
  intro h
  obtain ⟨x, y, hxy⟩ := h
  -- Analyze the possible values of x^3 modulo 7.
  have hx : x^3 %
    have hx : x %
      omega
    rcases hx with (h | h | h | h | h | h | h) <;>
      simp [h, pow_three, Int.mul_emod, Int.add_emod]
  -- Analyze the possible values of y^4 modulo 7.
  have hy : y^4 %
    have hy : y %
      omega
    rcases hy with (h | h | h | h | h | h | h) <;>
      simp [h, pow_four, Int.mul_emod, Int.add_emod]
  -- Combine the results to find a contradiction.
  have h7 : (x^3 + y^4) %
    rcases hx with (h | h | h) <;>
      rcases hy with (h' | h' | h') <;>
        simp [h, h', Int.add_emod, Int.mul_emod]
  -- Conclude the proof by contradiction.
  @!\colorbox{c1}{omega}@!
\end{lstlisting}
\par
\vspace{0.1\baselineskip}
\textbf{\scriptsize Vanilla (Not Success):} 

\lstset{
    frame=single,
    basicstyle=\scriptsize,
    xleftmargin=0.1pt,
    framexleftmargin=0.1pt,
    keywordstyle=\bfseries,
    morecomment=[l]{--},   %
    commentstyle=\color{gray}, 
    escapeinside={@!}{@!},     %
     moredelim=**[is][\color{red}]{@}{@},
     language=lean,
}

\begin{lstlisting}[frame=single, basicstyle=\scriptsize]
theorem algebra_68653_vanilla : ¬ ∃ x y : ℤ, x^3 + y^4 = 7 := by
  intro h; rcases h with ⟨x, y, hxy⟩; have hx : x^3 ≥ 0 := by positivity; have hy : y^4 ≥ 0 := by positivity
  @!\colorbox{c1}{linarith}@!
\end{lstlisting}
\scriptsize
\label{textbox:dpsk_cot}
\end{tcolorbox}

\section{Delving into Common Error Patterns of Existing Provers}

\subsection{Experimental Settings}

In this section, we systematically analyze common error patterns observed in existing theorem provers (\eg, DeepSeek-V1.5, STP, Goedel, and Kima-Prover).
We employ advanced general‑purpose LLMs (\eg, o4‑mini) to automate both error diagnosis and classification: first extracting salient verbalized features (\ie,
Example~\ref{textbox:prompt_vfe}), then assigning the found error features via a second prompt‑driven call (\ie, Example~\ref{textbox:prompt_classification}). 
For each prover, we randomly sampled 100 failed proofs from a variety of Lean 4 statements and processed them through our two-stage diagnosis and classification pipeline. 
Human domain experts then manually reviewed and corrected both the extracted features and the preliminary labels. 
We identified the four most common failure patterns---incomplete proofs, inability to handle complex inequalities, improper use of automation tactics, and redundant hypothesis introduction---as summarized in Table \ref{tab:Error_type}. 
Note that a single proof attempt may exhibit multiple errors, so the percentages do not sum to 100\%.

\subsection{Error Patterns Analysis and Case Study}

\textbf{Improper Use of Automation Tactics.} Existing LLM-based Lean4 provers frequently generate proofs that rely heavily on automation tactics -- such as \texttt{aesop}~\cite{limperg2023aesop}, \texttt{simp}, and \texttt{linarith}, to streamline the low‑level, step‑by‑step reasoning required by tactic-based proofs.
For example, \texttt{aesop} performs a best‑first proof search over a database of tagged lemmas and applies rewriting, splitting, and instance search to discharge goals.
But these tactics depend on fixed heuristics and pre‑tagged lemmas that may not match the structure of every proof: when over‑invoked or misconfigured, they can dramatically expand the search space, lead to nontermination or timeouts, or even transform goals into irrelevant or unsolvable forms.
In particular, automated tactics often struggle to supply the explicit constructions or witnesses required by truly constructive proofs~\cite{smith1995constructive}, which may discharge the main proposition without building the underlying data, resulting in incomplete or invalid reasoning. 

\vspace{1.5mm}
Taking the failed proof of \texttt{omni\_theorem\_4000} as an example, it fails to construct a witness $a$ within the correct domain that satisfies both (1) $a \le 1 \vee a >0$ and (2) \( f(x) = \begin{cases}
0, & \text{if } x \ne -a^2 \\
a, & \text{if } x = -a^2
\end{cases} \).
Instead of performing case-by-case analysis, the proof, however, introduces the incorrect witness $a = 0$, and relies on \texttt{simp} to close off the remaining goals that are not designed to solve, without specifically analyzing the core function $(x + y^2) \cdot f(y \cdot f(x)) = x \cdot y \cdot f(y^2 + f(x))$.

\setlength{\intextsep}{0pt}
\setlength{\columnsep}{11pt}
\setlength{\abovecaptionskip}{3pt}
\setlength{\belowcaptionskip}{2pt}
\setlength{\tabcolsep}{10pt} %
\renewcommand{\arraystretch}{1.35} %
\begin{table} %
\centering
\small
\begin{tabular}{lccccc}
\textbf{Error} & \textbf{DeepSeek-SFT} & \textbf{DeepSeek-RL} & \textbf{Goedel} & \textbf{STP} & \textbf{Kimina }\\
\shline
Redundant Hypothesis     & 18.0\% & 34.0\% & 27.0\% & 24.0\% & 36.0\% \\
Incomplete Proof         & 77.0\% & 62.0\% & 86.0\% & 44.0\% & 93.0\% \\
Inabilities for Inequality & 8.0\% & 13.0\% & 20.0\% & 1.0\% & 20.0\% \\
Misuse of Auto tactics   & 62.0\% & 65.0\% & 78.0\% & 74.0\% & 43.0\% \\
\end{tabular}
\caption{\small Percentage of different Lean4 error patterns in LLM-based provers.}
\label{tab:Error_type}
\vspace{-2.5mm}
\end{table}

\vspace{2mm}
\vspace{0.5\baselineskip}
\begin{table}[!t]
    \centering
    \belowrulesep=0pt
    \aboverulesep=0pt
    \setlength{\abovecaptionskip}{6pt}
    \setlength{\belowcaptionskip}{-10pt}
    \setlength{\tabcolsep}{3pt}
    \renewcommand{\arraystretch}{1.2}
    \small
    \begin{tabular}{p{5cm}|p{6.5cm}|p{5cm}} 
    \multicolumn{1}{c|}{\textbf{Lean4 Statement}} & 
    \multicolumn{1}{c|}{\textbf{Prover Output }} & 
    \multicolumn{1}{c}{\textbf{Explanation}} \\
    \noalign{\global\arrayrulewidth=0.5pt}\shline\noalign{\global\arrayrulewidth=0.4pt}
\begin{lstlisting}[basicstyle=\scriptsize]

theorem omni_theorem_4000 (f : ℝ → ℝ) :
    (∀ x y, (x + y ^ 2) * f (y * f x) = x * y * f (y ^ 2 + f x)) ↔
    ∀ x, f x = x ∨ f x = 0 ∨ ∃ a, a ≤ -1 ∨ a > 0 ∧ f x = if x ≠ -a ^ 2 then 0 else a := by
\end{lstlisting}
    &
    \lstset{
    basicstyle=\scriptsize,
    xleftmargin=0.1pt,
    framexleftmargin=0.1pt,
    keywordstyle=\bfseries,
    morecomment=[l]{--},   %
    commentstyle=\color{gray}, 
    escapeinside={@}{@}     %
}
\begin{lstlisting}[basicstyle=\scriptsize]
      -- Proof Before
      -- If f(x) ≠ 0, we need to show that there exists an a such that the conditions hold.
        right
        right
       @\colorbox{c1}{use 0}@
        -- We choose a = 0 to satisfy the conditions.
       @\colorbox{c1}{simp}@
      --Proof After
\end{lstlisting}
    &
    \textbf{Improper use of Automation Tactics.} The prover selects the witness $a=0$ without checking the original side‑condition and then calls \texttt{simp} to close off the goal. This bypasses the essential case‑by‑case analysis of the core functional equation, resulting in an invalid proof that never constructs the required nonzero witness when needed.
    
    \\[-1mm]

    \midrule
\begin{lstlisting}[ basicstyle=\scriptsize]
theorem DEMIMathAnalysis_50 :
    Tendsto (λ n => sqrt n * ∫ x in univ, 1 / (1 + x ^ 2) ^ n) atTop (N (sqrt π)) := by
\end{lstlisting}
    &
    \lstset{
    basicstyle=\scriptsize,
    xleftmargin=0.1pt,
    framexleftmargin=0.1pt,
    keywordstyle=\bfseries,
    morecomment=[l]{--},   %
    commentstyle=\color{gray}, 
    escapeinside={@!}{@!}     %
}

\begin{lstlisting}[basicstyle=\scriptsize]
  rw [show sqrt = fun x => sqrt x by funext; rfl]
  @!\colorbox{c1}{apply tendsto\_atTop\_mono}@! (fun n => sqrt_le_sqrt <|
  integral_mono_on _ _ <| fun x _ => by
    gcongr
    all_goals norm_num)
  refine' tendsto_atTop_of_eventually_const (by simp)
  all_goals
    filter_upwards with _
  @!\colorbox{c1}{<;> simp}@!
\end{lstlisting}
    & 
    \textbf{Incomplete Proof.} The proof invokes \texttt{tendsto\_atTop\_mono} without establishing any of the analytical prerequisites. 
    The final \texttt{simp} steps trivially handle eventual constancy but leave the main asymptotic argument unresolved.
    \\
    \midrule
    \begin{lstlisting}[basicstyle=\scriptsize]


theorem aime_all_2005_II_1 (n : ℕ) (hn : n ≥ 6) (h : {s : Finset (Fin n) | s.card = 6}.ncard = 6 * {s : Finset (Fin n) | s.card = 3}.ncard) :
    n = 13 := by 
  \end{lstlisting}
    &
    \lstset{
    basicstyle=\tiny,
    xleftmargin=0.1pt,
    framexleftmargin=0.1pt,
    keywordstyle=\bfseries,
    morecomment=[l]{--},   %
    commentstyle=\color{gray}, 
    escapeinside={@!}{@!}     %
}

\begin{lstlisting}[basicstyle=\scriptsize]
  norm_num [Nat.choose] at h
  revert n hn h
 @!\colorbox{c1}{rintro n hn h}@!
  ring_nf at h
  nlinarith
\end{lstlisting}
  & 
\textbf{Redundant Hypothesis Introduction.} After \texttt{norm\_num} already rewrites the binomial coefficients, the \texttt{revert} ; \texttt{rintro} sequence merely re‑introduces \texttt{n}, \texttt{hn}, and \texttt{h} in the same form, adding no new information and cluttering the context. 
 \\[-1mm]
 \midrule
 \begin{lstlisting}[basicstyle=\scriptsize]

theorem algebra_528739 (a b c : ℝ) (ha : 0 < a) (hb : 0 < b) (hc : 0 < c) :
    1 / (a^3 + b^3 + a * b * c) + 1 / (b^3 + c^3 + a * b * c) +
    1 / (c^3 + a^3 + a * b * c) ≤ 1 / (a * b * c) := by 
  \end{lstlisting}
    &
    \lstset{
    basicstyle=\scriptsize,
    xleftmargin=0.1pt,
    framexleftmargin=0.1pt,
    keywordstyle=\bfseries,
    morecomment=[l]{--},   %
    commentstyle=\color{gray}, 
    escapeinside={@}{@}     %
}
\begin{lstlisting}[basicstyle=\scriptsize]
  -- We start by simplifying the left-hand side using the fact that each term is non-negative.
  have h₀ : 0 < a * b * c := by positivity
  ...
  -- We then apply the inequality to each term.
  rw [div_le_div_iff] <;>
  @\colorbox{c1}{nlinarith}@, [sq_nonneg (a - b), sq_nonneg (b - c), sq_nonneg (c - a)]
  ...
\end{lstlisting}
  & 
  \textbf{Inadequate Handling of Inequalities.}
The solver attempts to apply \texttt{nlinarith} after a single \texttt{div\_le\_div\_iff}, but the cyclic, high‑degree fractional structure exceeds its linear‐and‐quadratic reasoning scope.
\\[-1mm]
    \end{tabular}
    \vspace{-1.5mm}
    \caption{\small Examples of common Lean4 error patterns in LLM-based provers.}
    \label{tab:proof_case}
    \vspace{1.5mm}
\end{table}

\noindent \textbf{Inabilities to Handle Complex Inequalities.}
Current provers over-rely on \texttt{linarith} and \texttt{nlinarith} to find contradictions between hypotheses that are linear and some non-linear (in)equalities. 
Common procedures using them require the provers to (1) mix high‑degree polynomials and rational functions, (2) exploit cyclic or symmetric structure, and (3) use domain‑specific lemmas (\eg, rearrangements, Chebyshev, AM–GM variants).

For the failed proof \texttt{algebra\_528739}, \texttt{nlinarith} must first clear denominators in the sum of fractions by introducing the common denominator: $D \;=\;\bigl(a^3+b^3+abc\bigr)\,\bigl(b^3+c^3+abc\bigr)\,\bigl(c^3+a^3+abc\bigr)$.
However, expanding \(D\) yields a degree‑9 polynomial in three variables with $\sim 55$ (via \(\binom{9+3-1}{3-1}\approx55\)) monomials, rendering sum‑of‑squares or Fourier–Motzkin methods infeasible.
Even if somehow the denominator are manually cleared, \texttt{nlinarith} can only handle (1) linear combinations of monomials (via \texttt{linarith}), (2) quadratic forms (by introducing auxiliary square variables and then linearizing), and (3) simple monotonicity lemmas (\eg, if \(0 < x \le y \implies \frac{1}{x} \ge \frac{1}{y}\)), but only after the provers normalize the goal via \texttt{ring} or \texttt{field} first. 
In contrast, a standard deductive reasoning for this problem would be: (1) Prove $a^3+b^3+abc \ge abc$ by AM-GM inequality or rearrangement, (2) Conclude $\frac{1}{a^3+b^3+abc}\leq\frac{1}{abc}$ and similarly for the other two cyclic terms, (3) Sum up the three inequalities to get the result.
While provers attempt to invoke \texttt{nlinarith} directly, without these intermediate deductive steps, it leads to failure.

\vspace{1.5mm}
\noindent \textbf{Redundant Hypothesis Introduction.}
A common error in current LLM-based theorem provers arises from introducing structurally redundant hypotheses.
While these do not inherently cause logical errors, they obscure the proof’s underlying logic and reduce readability.
For example, in the \texttt{aime\_all\_2005\_II\_1} proof (Table~\ref{tab:proof_case}), the unnecessary use of \texttt{revert} followed by \texttt{reintro} exemplifies this issue.
These tactics are designed to generalize variables or hypotheses---a technique critical for inductive proofs or hypothesis strengthening. 
However, in this case: (1) no inductive reasoning requires generalization, (2) the variables \texttt{n}, \texttt{hn}, and \texttt{h} already exist in the context and can be directly used.
Therefore, the use of \texttt{revert} is redundant and can be removed to simplify the proof.

\vspace{1.5mm}
\noindent \textbf{Incomplete Proof.}
Another common failure mode for for LLM-based provers is generating unfinished proof attempts that leave critical subgoals unresolved or rely on placeholder tactics without justifying intermediate steps. For example, in the proof sketch for \texttt{DEMIMathAnalysis\_50}, which aims to show $\lim_{n\to\infty}\sqrt{n}\cdot\int_{-\infty}^{\infty}\frac{1}{(1+x^2)^n}dx=\sqrt{\pi}$, the prover terminates prematurely after a few tactic calls that: (1) fail to justify interchanging the limit and integral and (2) fail to establish bounds on the integrand's tail decay.

\noindent The flawed proof begins with an unnecessary rewrite of \texttt{sqrt} and misapplies monotonicity lemmas like \texttt{integral\_mono\_on} without verifying domination or integrability conditions required for the Dominated Convergence Theorem. 
Worse, tactics such as \texttt{tendsto\_atTop\_of\_eventually\_const} and \texttt{filter\_upwards} trivialize tail behavior instead of rigorously addressing convergence.

We hypothesize this error stems from short-sighted heuristic selection during language modeling of theorem provers: prioritizing tactics that maximize immediate log-probability or heuristic scores (\eg, \texttt{gcongr}, \texttt{norm\_num}, \texttt{simp}) over those advancing global proof progress. 
Such choices syntactically reshape goals while burying core challenges under shallow subgoals.

\section{Concluding Remarks}

We introduce ForamlMATH, a novel and extensive benchmark for evaluating the formal mathematical reasoning capabilities of LLMs.
Comprising 5,560 formally verified statements in Lean4, FormalMATH spans a wide range of mathematical domains, including algebra, number theory, calculus, and discrete mathematics, encompassing problems from high-school Olympiad level to undergraduate curricula.
We propose a simple yet effective human-in-the-loop autoformalization pipeline to construct FormalMATH. 
This pipeline integrates specialized LLMs for initial Lean4 statement formalization, multi-LLM semantic verification to ensure fidelity to the original natural-language problems, and a negation-based disproof strategy for filtering invalid statements, which extensively reduces the effort for subsequent manual review by human experts, while achieving a high pre-verification preservation rate of 72.09\%.

Our comprehensive evaluation of state-of-the-art LLM-based theorem provers on FormalMATH reveals significant limitations in current systems. Even the most capable models demonstrate modest success rates under practical sampling budgets, with the top performer achieving only 16.46\% accuracy. Our analysis further identifies pronounced domain biases, wherein models excel in certain domains like algebra but struggle considerably in other domains such as calculus. 
Additionally, our findings indicate an over-reliance on simplified automation tactics and, counterintuitively, a negative impact of natural-language solution guidance on proof success in CoT scenarios. 
These results highlight the challenging nature of the FormalMATH benchmark and pose critical open problems for future research in enhancing robustness, generalizability, and reasoning complexity of automatic theorem provers.

\bibliography{ref}
\bibliographystyle{alpha}
\newpage

\newpage

\appendix

\addcontentsline{toc}{section}{Appendix} %
\renewcommand \thepart{} %
\renewcommand \partname{}
\part{\Large{\centerline{Appendix}}}
\parttoc

\newpage

\section{Data Sources}\label{app:data_source}
Table~\ref{tab:datasets} presents the sources of the natural language datasets used in the FormalMATH project.

\vspace{3.5mm}
\begin{table}[h]
\small
\centering
\setlength{\abovecaptionskip}{4pt}
\setlength{\tabcolsep}{10pt}
\renewcommand{\arraystretch}{1.3}
\begin{tabular}{l|ccccc}
\textbf{Dataset} & \textbf{Level} & \textbf{\#Domains} & \textbf{Size} & \textbf{\#S.Formal} \\ 
\shline
Omni-math~\cite{gao2024omni} & High School Olympiad & 9 & 4.43k & 1,210 \\
Numina-Olympiad & High School Olympiad & 10 & 11.8k & 2,409 \\
AIME-Math & High School Olympiad & 7 & 934 & 371 \\
BlueMO~\citep{bluemo2024} & High School Olympiad & 8 & 3,024 & 1,099 \\
U-Math~\cite{chernyshev2024u} & Undergraduate & 6 & 1,100 & 358 \\
Hardmath~\cite{fan2024hardmath} & Undergraduate & 3 & 1,466 & 67 \\
DEMIMATH~\cite{hata2007problems} & Undergraduate & 8 & 2,830 & 46 \\
\end{tabular}
\caption{\small The sources of the datasets in \our. ``\#Domains'' denotes the number of domains in the dataset. ``\#S.Formal'' denotes the number of samples in \our \;that are formalized from the dataset.}\label{tab:datasets}
\end{table}

\section{Examples of Statements in FormalMATH}
\label{sec:example}
\begin{tcolorbox}[
    title=\textbf{\scriptsize  },
    colframe=c3,
    boxrule=0.5pt,
    colback=c4!5,
    fonttitle=\scriptsize, %
    fontupper=\scriptsize, %
]
\footnotesize
\vspace{0.2em}
\textbf{\scriptsize u-math\_915}
\begin{lstlisting}[frame=single, basicstyle=\scriptsize]
Find the derivative of the 27th order $y^{(27)}$ for a function $y = 2 \cdot x^2 \cdot \sin(x)$. Prove that the answer is: $y^{(27)}$ = $1404\cdot\cos(x)-2\cdot x^2\cdot\cos(x)-108\cdot x\cdot\sin(x)$
\end{lstlisting}
\par
\vspace{0.5\baselineskip}
\textbf{\scriptsize Autoformalization:} 

\lstset{
    frame=single,
    basicstyle=\scriptsize,
    xleftmargin=0.1pt,
    framexleftmargin=0.1pt,
    keywordstyle=\bfseries,
    morecomment=[l]{--},   %
    commentstyle=\color{gray}, 
    escapeinside={@}{@}     %
}

\begin{lstlisting}[frame=single, basicstyle=\scriptsize]
import Mathlib

open Real Set
open scoped BigOperators

theorem u_math_915 {f : ℝ → ℝ} (hf : f = λ x => 2 * x ^ 2 * sin x) :
    iteratedDeriv 27 f = λ x => 1404 * cos x - 2 * x ^ 2 * cos x - 108 * x * sin x := by 
\end{lstlisting}
\tiny
\end{tcolorbox}
\vspace{1mm}

\begin{tcolorbox}[
    title=\textbf{\scriptsize  },
    colframe=c3,
    boxrule=0.5pt,
    colback=c4!5,
    fonttitle=\scriptsize, %
    fontupper=\scriptsize, %
]
\footnotesize
\vspace{0.2em}
\textbf{\scriptsize hardmath\_569}
\begin{lstlisting}[frame=single, basicstyle=\scriptsize]
Consider the integral
$I(x) = \int_{-0.3}^{0.6} \left( 1.2 t^5 + 2.4 t^4 + 1.7 t^3 - 0.6 \operatorname{atan}(t) + 1.3 \right) e^{x \left( 1.2 t^5 - 0.8 t^4 - 2.7 t^3 - 2.6 t^2 \right)} \, dt
$
Develop an analytical formula for $I(x)$ that is accurate as $x \to \infty$.
Prove that the answer is:
$
I(x) \approx \frac{\sqrt{65} \sqrt{\pi} \sqrt{\frac{1}{x}}}{10}.
$



 
\end{lstlisting}
\par
\vspace{0.5\baselineskip}
\textbf{\scriptsize Autoformalization:} 

\lstset{
    frame=single,
    basicstyle=\scriptsize,
    xleftmargin=0.1pt,
    framexleftmargin=0.1pt,
    keywordstyle=\bfseries,
    morecomment=[l]{--},   %
    commentstyle=\color{gray}, 
    escapeinside={@}{@}     %
}

\begin{lstlisting}[frame=single, basicstyle=\scriptsize]
import Mathlib

open Real

theorem hardmath_569 ($f : \mathbb{R} \to \mathbb{R}$) (hf : f = fun x => ∫ t in (-0.3)..0.6, (1.2 * t^5 + 2.4 * t^4 + 1.7 * t^3 - 0.6 * arctan t + 1.3) * exp (x * (1.2 * t^5 - 0.8 * t^4 - 2.7 * t^3 - 2.6 * t^2))) : 

∀ ε > 0, ∃ x, ∀ y, y > x → |f y - (Real.sqrt 65 * Real.sqrt π * Real.sqrt (1 / y)) / 10| < ε := by
\end{lstlisting}
\tiny
\end{tcolorbox}
\vspace{1mm}

\begin{tcolorbox}[
    title=\textbf{\scriptsize  },
    colframe=c3,
    boxrule=0.5pt,
    colback=c4!5,
    fonttitle=\scriptsize, %
    fontupper=\scriptsize, %
]
\footnotesize
\vspace{0.2em}
\textbf{\scriptsize omni\_theorem\_4136}
\begin{lstlisting}[frame=single, basicstyle=\scriptsize]
Determine all pairs $(x,y)$ of positive integers such that $x^{2}y+x+y$ is divisible by $xy^{2}+y+7$.

Prove that the answer is: $(x,y) = (11,1), (49,1), (7t^2,7t), t \text{ is an interge}$



 
\end{lstlisting}
\par
\vspace{0.5\baselineskip}
\textbf{\scriptsize Autoformalization:} 

\lstset{
    frame=single,
    basicstyle=\scriptsize,
    xleftmargin=0.1pt,
    framexleftmargin=0.1pt,
    keywordstyle=\bfseries,
    morecomment=[l]{--},   %
    commentstyle=\color{gray}, 
    escapeinside={@}{@}     %
}

\begin{lstlisting}[frame=single, basicstyle=\scriptsize]
import Mathlib

open Real

open scoped BigOperators

theorem omni_theorem_4136 :  ∀ x y : ℕ, x > 0 ∧ y > 0 → ((x ^ 2 * y + x + y) %

(x = 11 ∧ y = 1) ∨ (x = 49 ∧ y = 1) ∨ ∃ t : ℤ, x = 7 * t ^ 2 ∧ y = 7 * t) := by
\end{lstlisting}
\tiny
\end{tcolorbox}
\vspace{1mm}
\section{The Error Types of Our Autoformalization Pipeline}\label{app:errortype}

\begin{table}[ht]
\small
\centering
\setlength{\abovecaptionskip}{4pt}
\setlength{\belowcaptionskip}{-5pt}
\setlength{\tabcolsep}{10pt}
\renewcommand{\arraystretch}{1.25}
    \begin{tabular}{l|c}
        \textbf{Error Category} & \textbf{Percentage (\%)} \\
        \shline
        Condition Error & 83.3 \\
        Expression Error (Lean Syntax) & 9.6 \\
        Definition Error (No Mathematical Meaning) & 3.3 \\
        Domain Error & 1.7 \\
        Propositional Logic Error & 0.9 \\
        Lack of Geometric Background & 0.6 \\
        Condition Redundancy & 0.5 \\
        Algebraic Expression Error & 0.2 \\
    \end{tabular}
    \caption{\small Error Classification Statistics (\%)}\label{tab:error_stats}
\end{table}

\section{Domain Distribution of FormalMATH-Lite}\label{app:domain_lite}

\begin{figure}[h]
  \centering
  \setlength{\abovecaptionskip}{4pt}
  \setlength{\belowcaptionskip}{4pt}
  \vspace{-1mm}
  \includegraphics[width=0.99\textwidth]{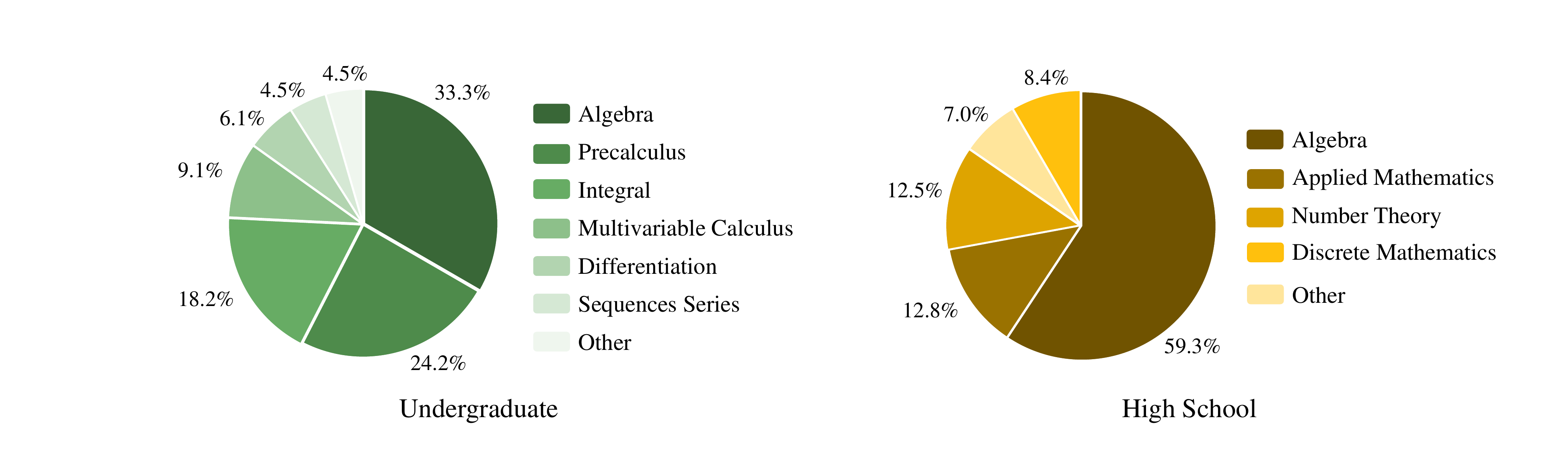}
  \caption{\small The distribution of mathematical domains in our FormalMATH-Lite dataset.}
  \label{fig:domain_lite}
  \vspace{-2.5mm}
\end{figure}

\newpage
\section{Typical Errors in Statement Autoformalization}
\subsection{Errors in Definition}
Definition Errors mainly occur in the failure of semantically corresponding the mathematical entities in autoformalized statements one-to-one with those in the refined statement. For instance, in a geometry problem, if the defined entities do not pertain to geometric concepts such as points, lines, and angles, it indicates a lack of domain relevance.

Mismatched Variable type is also a common issue. This type of error can lead to failure when applying tactics that are specific to a particular variable type.

Furthermore, the defined variables may lack actual mathematical significance so that the statement cannot pass the compiler.
\vspace{0.5\baselineskip}
\begin{tcolorbox}[
    title=\textbf{\scriptsize Errors in Definition: Lack of Specific Mathematical Background (Geometry) },
    colframe=c3,
    boxrule=0.5pt,
    colback=c4!5,
    fonttitle=\scriptsize, %
    fontupper=\scriptsize, %
]
\footnotesize
\vspace{0.2em}
\textbf{\scriptsize omni\_theorem\_2599}
\begin{lstlisting}[frame=single, basicstyle=\scriptsize]
Points $A, B, C, D$ lie on a circle in that order such that $\frac{A B}{B C}=\frac{D A}{C D}$. If $A C=3$ and $B D=B C=4$, find $A D$.
 Prove that the answer is: $\frac{3}{2}$
\end{lstlisting}
\par
\vspace{0.5\baselineskip}
\textbf{\scriptsize Autoformalization:} 

\lstset{
    frame=single,
    basicstyle=\scriptsize,
    xleftmargin=0.1pt,
    framexleftmargin=0.1pt,
    keywordstyle=\bfseries,
    morecomment=[l]{--},   %
    commentstyle=\color{gray}, 
    escapeinside={@}{@}     %
}

\begin{lstlisting}[frame=single, basicstyle=\scriptsize]
import Mathlib

open EuclideanGeometry Real
open scoped BigOperators

theorem omni_theorem_2599
  {A B C D : EuclideanSpace ℝ (Fin 2)}
  (h₀ : ∃ O, dist A O = dist B O ∧ dist B O = dist C O ∧ dist C O = dist D O)
  (h₁ : dist A B / dist B C = dist D A / dist C D)
  (h₂ : dist A C = 3)
  (h₃ : dist B D = 4)
  (h₄ : dist B C = 4) :
  dist A D = 3 / 2 := by sorry
\end{lstlisting}
\tiny
\end{tcolorbox}
\vspace{1mm}

The model's formalization is deficient because it only specifies the lengths of line segments $AB$, $BC$, $CD$, and $DA$, without explicitly stating that points $A$, $B$, $C$, and $D$ are in that order along circle O. This omission is critical as only segment lengths is insufficient to uniquely define a quadrilateral inscribed in a circle. Different orderings of these points on the circle, even with the same segment lengths.
\vspace{0.5\baselineskip}
\begin{tcolorbox}[
    title=\textbf{\scriptsize Errors in Definition: Mismatched variable Type },
    colframe=c3,
    boxrule=0.5pt,
    colback=c4!5,
    fonttitle=\scriptsize, %
    fontupper=\scriptsize, %
]
\footnotesize
\vspace{0.2em}
\textbf{\scriptsize omni\_theorem\_66}
\begin{lstlisting}[frame=single, basicstyle=\scriptsize]
Determine whether or not there exist two distinct sets $A$ and $B$, each consisting of at most $2011^2$ positive integers, such that for every $x$ with $0 < x < 1$, the following inequality holds:
$\left| \sum_{a \in A} x^a - \sum_{b \in B} x^b \right| < (1-x)^{2011}.$
Prove that the answer is Yes.
\end{lstlisting}
\par
\vspace{0.5\baselineskip}
\textbf{\tiny Autoformalization:} 

\lstset{
    frame=single,
    basicstyle=\scriptsize,
    xleftmargin=0.1pt,
    framexleftmargin=0.1pt,
    keywordstyle=\bfseries,
    morecomment=[l]{--},   %
    commentstyle=\color{gray}, 
    escapeinside={@}{@}     %
}

\begin{lstlisting}[frame=single, basicstyle=\scriptsize]
import Mathlib

theorem omni_theorem_66 :
    ∃ A B : Finset ℕ,
      A ≠ B ∧
      A.card ≤ 2011^2 ∧
      B.card ≤ 2011^2 ∧
      (∀ x ∈ Set.Ioo 0 1, |∑ a ∈ A, (x : ℝ)^a - ∑ b ∈ B, (x : ℝ)^b| < (1 - x)^2011) := by sorry
\end{lstlisting}
\scriptsize
\end{tcolorbox}
While positive integer mappings and positive integers m and n were required, the autoformalization mistakenly used natural numbers instead.

\vspace{0.5\baselineskip}
\begin{tcolorbox}[
    title=\textbf{\scriptsize Errors in Definition: Meaningless Definition in Mathematical Context },
    colframe=c3,
    boxrule=0.5pt,
    colback=c4!5,
    fonttitle=\scriptsize, %
    fontupper=\scriptsize, %
]
\scriptsize
\vspace{0.2em}
\textbf{\scriptsize omni\_theorem\_1024}
\begin{lstlisting}[frame=single, basicstyle=\scriptsize]
For a positive integer $n$, denote by $\tau(n)$ the number of positive integer divisors of $n$, and denote by $\phi(n)$ the number of positive integers that are less than or equal to $n$ and relatively prime to $n$. Call a positive integer $n$ good if $\varphi(n)+4 \tau(n)=n$. For example, the number 44 is good because $\varphi(44)+4 \tau(44)=44$. Find the sum of all good positive integers $n$.
 Prove that the answer is: 172
\end{lstlisting}
\par
\vspace{0.5\baselineskip}
\textbf{\scriptsize Autoformalization:} 

\lstset{
    inputencoding=utf8,
    extendedchars=true, 
    frame=single,
    basicstyle=\scriptsize,
    xleftmargin=0.1pt,
    framexleftmargin=0.1pt,
    keywordstyle=\bfseries,
    morecomment=[l]{--},   %
    commentstyle=\color{gray}, 
    escapeinside={@}{@}     %
}

\begin{lstlisting}[frame=single, basicstyle=\scriptsize]
import Mathlib

open Finset

def isGood (n : ℕ) := n.totient + 4 * n.divisors.card = n

theorem omni_theorem_1024 :
    ∑ n ∈ {n | 0 < n ∧ isGood n}, n = 172 := by sorry
\end{lstlisting}
\scriptsize
\end{tcolorbox}
Euler's totient function, denoted as $\phi(n)$ or in Lean 4 as \texttt{n\.totient}, counts the number of positive integers less than or equal to $n$ that are relatively prime to $n$. For $n=0$, the function \texttt{n.totient} is not defined, which results in isGood n being undefined.
\subsection{Errors in Expressions}
This type of errors include incorrect arithmetic or logical operations, improper use of variables, or incorrect application of mathematical or logical rules. 

\vspace{0.5\baselineskip}
\begin{tcolorbox}[
    title=\textbf{\scriptsize Errors in Proposition Construction (Wrong Position of Existential Qualifier },
    colframe=c3,
    colback=c4!5,
    boxrule=0.5pt,
    fonttitle=\scriptsize, %
    fontupper=\scriptsize, %
]
\scriptsize
\vspace{0.2em}
\textbf{\scriptsize omni\_theorem\_4}
\begin{lstlisting}[frame=single, basicstyle=\scriptsize]
Let $p$ be a prime number. Prove the following theorem of Euler: the equation $p=x^{2}+3y^{2}$ has a solution with $x, y \in \mathbb{Z}$ if and only if $p=3$ or $p \equiv 1 \pmod{3}$. (You may use the fact that the ring of integers of $\mathbb{Q}(\sqrt{-3})$ is a principal ideal domain.)
Prove that the answer is: The equation  $p = x^{2} + 3y^{2}$ has a solution with $ x, y \in \mathbb{Z} $ if and only if $$ p = 3 $$ or $$ p \equiv 1 \pmod{3} $$
\end{lstlisting}
\par
\vspace{0.5\baselineskip}
\textbf{\scriptsize Autoformalization:} 

\lstset{
    frame=single,
    basicstyle=\scriptsize,
    xleftmargin=0.1pt,
    framexleftmargin=0.1pt,
    keywordstyle=\bfseries,
    morecomment=[l]{--},   %
    commentstyle=\color{gray}, 
    escapeinside={@}{@}     %
}

\begin{lstlisting}[frame=single, basicstyle=\scriptsize]
import Mathlib

open Real
open scoped BigOperators

theorem omni_theorem_401 {p : ℕ} (hp : Nat.Prime p) :
  ∃ x y : ℤ, p = x^2 + 3*y^2 ↔ p = 3 ∨ p %
\end{lstlisting}
\scriptsize
\end{tcolorbox}
In the Lean 4 code, the existential quantifier is misplaced. $ \exists x, y \in \mathbb{Z}$ should quantify the entire left-hand side of the biconditional. Instead of asserting that the existence of a solution $(x, y)$ is equivalent to the condition $p = 3 \lor p \equiv 1 \pmod{3}$, the incorrect code unintentionally suggests the existence of $x$ and $y$ is linked to the entire biconditional statement being true. the existential quantifier should be correctly scoped only to the left-hand side of the biconditional, as in the corrected form: 
$$ (\exists x, y \in \mathbb{Z}, p = x^2 + 3y^2) \leftrightarrow (p = 3 \lor p \equiv 1 \pmod{3}), $$
which ensures that the quantifier only applies to the equation $p = x^2 + 3y^2$.
\vspace{0.5\baselineskip}
\begin{tcolorbox}[
    title=\textbf{\scriptsize Errors in Expression:  Misleading Scope of the Universal Quantifier },
    colback=c4!5,
    boxrule=0.5pt,
    colframe=c3,
    fonttitle=\scriptsize, %
    fontupper=\scriptsize, %
]
\scriptsize
\vspace{0.2em}
\textbf{\scriptsize omni\_theorem\_219}
\begin{lstlisting}[frame=single, basicstyle=\scriptsize]
Find, with proof, all nonconstant polynomials $P(x)$ with real coefficients such that, for all nonzero real numbers $z$ with $P(z) \neq 0$ and $P\left(\frac{1}{z}\right) \neq 0$, we have
$$
\frac{1}{P(z)}+\frac{1}{P\left(\frac{1}{z}\right)}=z+\frac{1}{z}
$$
Prove that the answer is:
$$
P(x) = \frac{x(x^{4k+2}+1)}{x^{2}+1} \quad \text{or} \quad P(x) = \frac{x(1-x^{4k})}{x^{2}+1}
$$

\end{lstlisting}
\par
\vspace{0.5\baselineskip}
\textbf{\scriptsize Autoformalization:} 

\lstset{
    frame=single,
    basicstyle=\scriptsize,
    xleftmargin=0.1pt,
    framexleftmargin=0.1pt,
    keywordstyle=\bfseries,
    morecomment=[l]{--},   %
    commentstyle=\color{gray}, 
    escapeinside={@}{@}     %
}

\begin{lstlisting}[frame=single, basicstyle=\scriptsize]
import Mathlib

open Fintype Set Real Ideal Polynomial
open scoped BigOperators

theorem omni_theorem_219 {P : Polynomial ℝ} (hP : P.degree > 0) :
    ∀ z : ℝ, z ≠ 0 → P.eval z ≠ 0 → P.eval (1/z) ≠ 0 →
      1/(P.eval z) + 1/(P.eval (1/z)) = z + 1/z ↔
  ∃ k : ℕ, P = X * (X^(4*k + 2) + 1)/(X^2 + 1) ∨
    P = X * (1 - X^(4*k))/(X^2 + 1) := by sorry
\end{lstlisting}
\scriptsize
\end{tcolorbox}
The problem is one of logical scope and intended semantic meaning. While Lean 4's type system is powerful and can often handle somewhat flexible syntax, the initial phrasing can lead to a misinterpretation of the theorem's claim. While Lean 4 might parse this code without immediate syntax errors due to the right-associativity of implication, this placement leads to a misinterpretation of the theorem's intended logical structure and meaning.
The original code is effectively parsed as if it were written:
\vspace{0.5\baselineskip}
\begin{tcolorbox}[
    title=\textbf{\scriptsize Logical Parse in Original Autoformalized Statement },
    colframe=c3,
    colback=c4!5,
    boxrule=0.5pt,
    fonttitle=\scriptsize, %
    fontupper=\scriptsize, %
]
\footnotesize
\vspace{0.2em}
\textbf{\scriptsize omni\_theorem\_219-autoformalization}
\begin{lstlisting}[frame=single, basicstyle=\scriptsize]
∀ z : ℝ, (z ≠ 0 → (P.eval z ≠ 0 → (P.eval (1/z) ≠ 0 → 
          (1/(P.eval z) + 1/(P.eval (1/z)) = z + 1/z ↔ 
      ∃ k : ℕ, P = X * (X^(4*k + 2) + 1)/(X^2 + 1) ∨ 
        P = X * (1 - X^(4*k))/(X^2 + 1) )))) 
\end{lstlisting}
\scriptsize
\end{tcolorbox}
The theorem should state: A nonconstant polynomial $P(x)$ satisfies the property that for all relevant nonzero $z$, the equation holds if and only if $P(x)$ takes one of the specified forms. To accurately reflect the intended meaning and correct the quantifier placement, we must use parentheses to explicitly define the scope of the universal quantifier. 
\vspace{0.5\baselineskip}
\begin{tcolorbox}[
    title=\textbf{\scriptsize Enhanced Autoformalized Statement },
    colback=c4!5,
    boxrule=0.5pt,
    colframe=c3,
    fonttitle=\scriptsize, %
    fontupper=\scriptsize, %
]
\footnotesize
\vspace{0.2em}
\textbf{\scriptsize omni\_theorem\_219-autoformalization}
\begin{lstlisting}[frame=single, basicstyle=\scriptsize]
theorem omni\_theorem\_219 \{P : Polynomial ℝ\} (hP : P.degree > 0) : \\
    (∀ z : ℝ, z ≠ 0 → P.eval z ≠ 0 → P.eval (1/z) ≠ 0 → 
      1/(P.eval z) + 1/(P.eval (1/z)) = z + 1/z) ↔ 
  ∃ k : ℕ, P = X * (X^(4*k + 2) + 1)/(X^2 + 1) ∨ 
    P = X * (1 - X^(4*k))/(X^2 + 1) := by sorry
\end{lstlisting}
\scriptsize
\end{tcolorbox}
\subsection{Errors in Constraint Condition}

The content of the constraints may not match the requirements stated in the problem or be left out. We also notice that constraints that are not mentioned in the original problem statement may be added during formalization.
\vspace{0.5\baselineskip}
\begin{tcolorbox}[
    title=\textbf{\scriptsize Errors in Constraint Condition: Mismatch with known conditions},
    colback=c4!5,
    boxrule=0.5pt,
    colframe=c3,
    fonttitle=\scriptsize, %
    fontupper=\scriptsize, %
]
\footnotesize
\vspace{0.2em}
\textbf{\scriptsize omni\_theorem\_748:}
\begin{lstlisting}[frame=single, basicstyle=\scriptsize]
Let $f(x)=x^{2}+a x+b$ and $g(x)=x^{2}+c x+d$ be two distinct real polynomials such that the $x$-coordinate of the vertex of $f$ is a root of $g$, the $x$-coordinate of the vertex of $g$ is a root of $f$ and both $f$ and $g$ have the same minimum value. If the graphs of the two polynomials intersect at the point (2012, - 2012), what is the value of $a+c$ ?
 Prove that the answer is: -8048
\end{lstlisting}
\par
\vspace{0.5\baselineskip}
\textbf{\scriptsize Autoformalization:} 

\lstset{
    frame=single,
    basicstyle=\scriptsize,
    xleftmargin=0.1pt,
    framexleftmargin=0.1pt,
    keywordstyle=\bfseries,
    morecomment=[l]{--},   %
    commentstyle=\color{gray}, 
    escapeinside={@}{@}     %
}

\begin{lstlisting}[frame=single, basicstyle=\scriptsize]
import Mathlib

theorem omni_theorem_748 {a b c d : ℝ} (f g : ℝ → ℝ)
    (hf : f = fun x => x^2 + a * x + b) (hg : g = fun x => x^2 + c * x + d)
    (hfg : f ≠ g) (hvertex : ∃ x, x = -a / 2 ∧ g x = 0) (hvertex' : ∃ x, x = -c / 2 ∧ f x = 0)
    (hmin : ∃ y, ∀ x, f x ≥ y ∧ g x ≥ y) (hintersect : f 2012 = -2012 ∧ g 2012 = -2012) :
    a + c = -8048 := by sorry
\end{lstlisting}
\scriptsize
\end{tcolorbox}
In Autoformalization, \texttt{hmin}: $\exists y, \forall x, f(x) \geq y \wedge g(x) \geq y$ means that $f$ and $g$ have a common lower bound, rather than: $f$ and $g$ have the same minimum value.

\vspace{0.5\baselineskip}
\begin{tcolorbox}[
    title=\textbf{\scriptsize Errors in Constraint Condition: Lack of a key condition},
    colframe=c3,
    boxrule=0.5pt,
    colback=c4!5,
    fonttitle=\scriptsize, %
    fontupper=\scriptsize, %
]
\footnotesize
\vspace{0.2em}
\textbf{\scriptsize omni\_theorem\_10:}
\begin{lstlisting}[frame=single, basicstyle=\scriptsize]
Given 30 students such that each student has at most 5 friends and for every 5 students there is a pair of students that are not friends, determine the maximum k such that for all such possible configurations, there exists k students who are all not friends.
Prove that the answer is: 6
\end{lstlisting}
\par
\vspace{0.5\baselineskip}
\textbf{\scriptsize Autoformalization:} 

\lstset{
    frame=single,
    basicstyle=\scriptsize,
    xleftmargin=0.1pt,
    framexleftmargin=0.1pt,
    keywordstyle=\bfseries,
    morecomment=[l]{--},   %
    commentstyle=\color{gray}, 
    escapeinside={@}{@}     %
}

\begin{lstlisting}[frame=single, basicstyle=\scriptsize]
import Mathlib

open Finset

theorem omni_theorem_10 (n : ℕ) (hn : n = 30) (f : Fin n → Fin n → Prop)
    (hf : ∀ i j, f i j ↔ f j i) (hfn : ∀ s : Finset (Fin n), s.card = 5 →
    ∃ i ∈ s, ∃ j ∈ s, i ≠ j ∧ ¬f i j) :
    IsGreatest {k | ∃ t : Finset (Fin n), t.card = k ∧ ∀ i ∈ t, ∀ j ∈ t, i ≠ j → ¬f i j} 6 := by sorry
\end{lstlisting}
\scriptsize
\end{tcolorbox}
The formalization fails to represent the condition ``Each student has at most 5 friends''. This crucial condition is necessary for accurately capturing the problem's constraints, and its absence may lead to a solution space that does not match the requirements of the original problem. Without this condition, the formalized statement does not fully reflect the situation described in the refined statement, potentially resulting in an incorrect or incomplete proof for determining the maximum number k of students who are all not friends. 
\vspace{0.5\baselineskip}
\begin{tcolorbox}[
    title=\textbf{\scriptsize Errors in Constraint Condition: Redundant conditions },
    colframe=c3,
    boxrule=0.5pt,
    colback=c4!5,
    fonttitle=\scriptsize, %
    fontupper=\scriptsize, %
]
\scriptsize
\vspace{0.2em}
\textbf{\scriptsize omni\_theorem\_1245}
\begin{lstlisting}[frame=single, basicstyle=\scriptsize]
The fraction $\frac{1}{2015}$ has a unique ``(restricted) partial fraction decomposition'' of the form $\frac{1}{2015}=\frac{a}{5}+\frac{b}{13}+\frac{c}{31}$ where $a, b, c$ are integers with $0 \leq a<5$ and $0 \leq b<13$. Find $a+b$.
 Prove that the answer is: 14
\end{lstlisting}
\par
\vspace{0.5\baselineskip}
\textbf{\scriptsize Autoformalization:} 

\lstset{
    frame=single,
    basicstyle=\scriptsize,
    xleftmargin=0.1pt,
    framexleftmargin=0.1pt,
    keywordstyle=\bfseries,
    morecomment=[l]{--},   %
    commentstyle=\color{gray}, 
    escapeinside={@}{@}     %
}

\begin{lstlisting}[frame=single, basicstyle=\scriptsize]
import Mathlib

theorem omni_theorem_1245 {a b c : ℤ} (ha : 0 ≤ a ∧ a < 5) (hb : 0 ≤ b ∧ b < 13)
    (hc : 0 ≤ c ∧ c < 31) (h : 1 / 2015 = a / 5 + b / 13 + c / 31) : a + b = 14 := by sorry
\end{lstlisting}
\scriptsize
\end{tcolorbox}
In autoformalization, the model introduced a spurious condition: 
\texttt{(hc : 0 $\leq$ c $\land$ c < 31)}.

\newpage
\section{Errors in Proof Goals}
The stated proof goals may be incorrect. They could be misaligned with the original problem statement, overly simplified, too general or even omitted.
\vspace{0.5\baselineskip}
\begin{tcolorbox}[
    title=\textbf{\scriptsize Errors in Proof Goals: Incorrect Proposition Logic },
    colframe=c3,
    boxrule=0.5pt,
    colback=c4!5,
    fonttitle=\scriptsize, %
    fontupper=\scriptsize, %
]
\scriptsize
\vspace{0.2em}
\textbf{\scriptsize omni\_theorem\_1362}
\begin{lstlisting}[frame=single, basicstyle=\scriptsize]
The Fibonacci sequence is defined as follows: $F_{0}=0, F_{1}=1$, and $F_{n}=F_{n-1}+F_{n-2}$ for all integers $n \geq 2$. Find the smallest positive integer $m$ such that $F_{m} \equiv 0(\bmod 127)$ and $F_{m+1} \equiv 1(\bmod 127)$.
 Prove that the answer is: 256
\end{lstlisting}
\par
\vspace{0.5\baselineskip}
\textbf{\scriptsize Autoformalization:} 

\lstset{
    frame=single,
    basicstyle=\scriptsize,
    xleftmargin=0.1pt,
    framexleftmargin=0.1pt,
    keywordstyle=\bfseries,
    morecomment=[l]{--},   %
    commentstyle=\color{gray}, 
    escapeinside={@}{@}     %
}
\begin{lstlisting}[frame=single, basicstyle=\scriptsize]
import Mathlib

def fib (n : ℕ) : ℕ := match n with
| 0 => 0
| 1 => 1
| n + 2 => fib (n + 1) + fib n

theorem omni_theorem_1362 {m : ℕ} (hm : m > 0) :
    fib m ≡ 0 [MOD 127] ∧ fib (m + 1) ≡ 1 [MOD 127] ↔ m = 256 := by sorry
\end{lstlisting}
\tiny
\end{tcolorbox}
The fundamental error in this autoformalization lies in framing the problem as an equivalence verification. The autoformalization, by using $\leftrightarrow m = 256$, transforms the problem into a claim that the given congruences hold if and only if $m$ is exactly 256. The correct approach to formalize the original problem would involve demonstrating that
The value $m=256$ satisfies the conditions:$ \text{fib} 256 \equiv 0 \pmod{127} \land \text{fib} (256 + 1) \equiv 1 \pmod{127} $,and that no smaller positive integer $m' < 256$ also satisfies these conditions. 
\vspace{0.5\baselineskip}
\begin{tcolorbox}[
    title=\textbf{\scriptsize Errors in Proof Goals: Incomplete Goal },
    colframe=c3,
    boxrule=0.5pt,
    colback=c4!5,
    fonttitle=\scriptsize, %
    fontupper=\scriptsize, %
]
\footnotesize
\vspace{0.2em}
\textbf{\scriptsize olymid\-ref\-base\_551}
\begin{lstlisting}[frame=single, basicstyle=\scriptsize]
Let $a, b, c, d, e, f$ be non-negative real numbers satisfying $a+b+c+d+e+f=6$. Find the maximal possible value of  $$ a b c+b c d+c d e+d e f+e f a+f a b $$  and determine all 6-tuples $(a, b, c, d, e, f)$ for which this maximal value is achieved.  Answer: 8 .  Prove that the answer is: 8
\end{lstlisting}
\par
\vspace{0.5\baselineskip}
\textbf{\scriptsize Autoformalization:} 

\lstset{
    frame=single,
    basicstyle=\scriptsize,
    xleftmargin=0.1pt,
    framexleftmargin=0.1pt,
    keywordstyle=\bfseries,
    morecomment=[l]{--},   %
    commentstyle=\color{gray}, 
    escapeinside={@}{@}     %
}
\begin{lstlisting}[frame=single, basicstyle=\scriptsize]
import Mathlib  open Real  theorem olymid_ref_base_551 :     IsGreatest {x | ∃ a b c d e f : ℝ, a ≥ 0 ∧ b ≥ 0 ∧ c ≥ 0 ∧ d ≥ 0 ∧ e ≥ 0 ∧ f ≥ 0 ∧ a + b + c + d + e + f = 6 ∧ x = a * b * c + b * c * d + c * d * e + d * e * f + e * f * a + f * a * b} 8 := by sorry
\end{lstlisting}
\scriptsize
\end{tcolorbox}
The core issue lies in how the autoformalization treats the problem's objective – finding the maximal possible value – and the request to determine all 6-tuples that achieve this maximum. The original problem requires the solver to not only find the maximum value but also to characterize the set of inputs that lead to this maximum. The provided autoformalization using IsGreatest completely omits any formalization of the requirement to determine all 6-tuples. It focuses solely on verifying the maximal value (8).

\newpage
\section{Prompt for Semantic Verification}
\label{sec:semantic}
To more effectively evaluate the consistency between natural language mathematics problems and their Lean4 formalizations, we adopted an LLMs group voting approach to filter entries with the same semantics. The prompt provided to the five LLMs is as follows:

\vspace{0.5\baselineskip}
\begin{tcolorbox}[
    title=\textbf{\scriptsize Prompt for Semantic Verification},
    colframe=c3,
    boxrule=0.5pt,
    colback=c4!5,
    fonttitle=\scriptsize, %
    fontupper=\scriptsize, %
    parskip=12pt
]
You are an expert in formalizing natural language into lean.\\
You are given a natural language statement and a lean statement.\\
You should judge the equivalence between the natural language statement and the lean statement by the following workflow:\\
1. You should back-translate the lean statement into English.\\
2. You should check if the back-translated statement is equivalent to the natural language statement.\\
3. If they are equivalent, you should return True.\\
4. Otherwise, you should return False.\\
Here is the natural language statement:\\
\{refined\_statement\} \\ 
Here is the lean statement: \\
\{lean\_statement\} \\ 
You must remember :Return True or False directly. Accept only True/False in answer.
\end{tcolorbox}

\section{Prompt for Domain Classification}

\begin{tcolorbox}[
    title=\textbf{\scriptsize Prompt for Domain Classification},
    colframe=c3,
    boxrule=0.5pt,
    colback=c4!5,
    fonttitle=\scriptsize, %
    fontupper=\scriptsize, %
    parskip=12pt,
    breakable
]
\# CONTEXT \#\\
I am a teacher, and I have some high-level math problems. \\
I want to categorize the domain of these math problems.\\

\# OBJECTIVE \#\\
A. Summarize the math problem in a brief sentence, describing the concepts involved in the math problem.\\
B. Categorize the math problem into specific mathematical domains. Please provide a classification chain, for example: Mathematics -> Applied Mathematics -> Probability -> Combinations. The following is a basic classification framework in the field of mathematics.\\
<math domains>\\
Mathematics\\
│\\
├── Applied Mathematics\\
│   ├── Math Word Problems\\
│   └── Statistics\\
│       ├── Mathematical Statistics\\
│       └── Probability\\
│           └── Counting Methods\\
│               ├── Permutations\\
│               └── Combinations\\
│\\
├── Algebra\\
│   ├── Prealgebra\\
│       ├── Integers\\
│       ├── Fractions\\
│       ├── Decimals\\
│       └── Simple Equations\\
│   ├── Algebra\\
│       ├── Algebraic Expressions\\
│       ├── Equations and Inequalities\\
│       ├── Factoring\\
│       └── Polynomial Operations\\
│   ├── Intermediate Algebra\\
│       ├── Quadratic Functions\\
│       ├── Exponential Functions\\
│       ├── Logarithmic Functions\\
│       └── Complex Numbers\\
│   ├── Linear Algebra\\
│       ├── Vectors\\
│       ├── Matrices\\
│       ├── Determinants\\
│       └── Linear Transformations\\
│   └── Abstract Algebra\\
│       ├── Group Theory\\
│       ├── Ring Theory\\
│       └── Field Theory\\
│\\
├── Geometry\\
│   ├── Plane Geometry\\
│       ├── Polygons\\
│       ├── Angles\\
│       ├── Area\\
│       ├── Triangulations\\
│       └── Perimeter\\
│   ├── Solid Geometry\\
│       ├── 3D Shapes\\
│       ├── Volume\\
│       └── Surface Area\\
│   ├── Differential Geometry\\
│       ├── Curvature\\
│       ├── Manifolds\\
│       └── Geodesics\\
│   └── Non-Euclidean Geometry\\
│       ├── Spherical Geometry\\
│       └── Hyperbolic Geometry\\
│\\
├── Number Theory\\
│   ├── Prime Numbers\\
│   ├── Factorization\\
│   ├── Congruences\\
│   ├── Greatest Common Divisors (GCD)\\
│   └── Least Common Multiples (LCM)\\
│\\
├── Precalculus\\
│   ├── Functions\\
│   ├── Limits\\
│   └── Trigonometric Functions\\
│\\
├── Calculus\\
│   ├── Differential Calculus\\
│       ├── Derivatives\\
│       ├── Applications of Derivatives\\
│       └── Related Rates\\
│   └── Integral Calculus\\
│       ├── Integrals\\
│       ├── Applications of Integrals\\
│       └── Techniques of Integration\\
│           ├── Single-variable\\
│           └── Multi-variable\\
│\\
├── Differential Equations\\
│   ├── Ordinary Differential Equations (ODEs)\\
│   └── Partial Differential Equations (PDEs)\\
│\\
└── Discrete Mathematics\\
    ├── Graph Theory\\
    ├── Combinatorics\\
    ├── Logic\\
    └── Algorithms\\
</math domains>\\

\# STYLE \#\\
Data report.\\

\# TONE \#\\
Professional, scientific.\\

\# AUDIENCE \#\\
Students. Enable them to better understand the domain and difficulty of the math problems.\\

\# RESPONSE: MARKDOWN REPORT \#
\#\# Summarization
[Summarize the math problem in a brief paragraph.]
\#\# Math domains
[Categorize the math problem into specific mathematical domains, including major domains and subdomains.]`

\# ATTENTION \#
 - The math problem can be categorized into multiple domains, but no more than three. Separate the classification chains with semicolons(;).\\
 - Your classification MUST fall under one of the aforementioned subfields; if it really does not fit, please add "Other" to the corresponding branch. For example: Mathematics -> Algebra -> Intermediate Algebra -> Other. Only the LAST NODE is allowed to be "Other"; the preceding nodes must strictly conform to the existing framework.\\
 - The math domain must conform to a format of classification chain, like "Mathematics -> Applied Mathematics -> Probability -> Combinations".\\
 - Add "=== report over ===" at the end of the report.\\

<example math problem>\\

{\raggedright
\text{[Question]}:
}\\
If $\frac{1}{9}+\frac{1}{18}=\frac{1}{square}$, what is the number that replaces the $square$ to make the equation true? \\
{\raggedright
\text{[Solution]}:
}\\
We simplify the left side and express it as a fraction with numerator 1: $\frac{1}{9}+\frac{1}{18}=\frac{2}{18}+\frac{1}{18}=\frac{3}{18}=\frac{1}{6}$. Therefore, the number that replaces the $square$ is 6.\\
{\raggedright
\text{[Source]: 2010\_Pascal}:
}\\

</example math problem>

\#\# Summarization
The problem requires finding a value that makes the equation $\frac{1}{9}+\frac{1}{18}=\frac{1}{square}$. 
This involves adding two fractions and determining the equivalent fraction.

\#\# Math domains
Mathematics -> Algebra -> Prealgebra -> Fractions;\\

=== report over ===\\

</example math problem>\\
{\raggedright
\text{[Question]}:
}\\
Let $\mathcal{P}$ be a convex polygon with $n$ sides, $n\ge3$. Any set of $n - 3$ diagonals of $\mathcal{P}$ that do not intersect in the interior of the polygon determine a triangulation of $\mathcal{P}$ into $n - 2$ triangles. If $\mathcal{P}$ is regular and there is a triangulation of $\mathcal{P}$ consisting of only isosceles triangles, find all the possible values of $n$. \\
{\raggedright
\text{[Solution]}:
}\\
We label the vertices of $\mathcal{P}$ as $P_0, P_1, P_2, \ldots, P_n$. Consider a diagonal $d = \overline{P_a\,P_{a+k}},\,k \le n/2$ in the triangulation. We show that $k$ must have the form $2^{m}$ for some nonnegative integer $m$.
This diagonal partitions $\mathcal{P}$ into two regions $\mathcal{Q},\, \mathcal{R}$, and is the side of an isosceles triangle in both regions. Without loss of generality suppose the area of $Q$ is less than the area of $R$ (so the center of $P$ does not lie in the interior of $Q$); it follows that the lengths of the edges and diagonals in $Q$ are all smaller than $d$. Thus $d$ must the be the base of the isosceles triangle in $Q$, from which it follows that the isosceles triangle is $\triangle P_aP_{a+k/2}\,P_{a+k}$, and so $2|k$. Repeating this process on the legs of isosceles triangle ($\overline{P_aP_{a+k/2}},\,\overline{P_{a+k}P_{a+k/2}}$), it follows that $k = 2^m$ for some positive integer $m$ (if we allow degeneracy, then we can also let $m=0$).
Now take the isosceles triangle $P_xP_yP_z,\,0 \le x < y < z < n$ in the triangulation that contains the center of $\mathcal{P}$ in its interior; if a diagonal passes through the center, select either of the isosceles triangles with that diagonal as an edge. Without loss of generality, suppose $P_xP_y = P_yP_z$. From our previous result, it follows that there are $2^a$ edges of $P$ on the minor arcs of $P_xP_y,\, P_yP_z$ and $2^b$ edges of $P$ on the minor arc of $P_zP_x$, for positive integers $a,\,b$. Therefore, we can write\[n = 2 \cdot 2^a + 2^b = 2^{a+1} + 2^{b},\]so $n$ must be the sum of two powers of $2$.
We now claim that this condition is sufficient. Suppose without loss of generality that $a+1 \ge b$; then we rewrite this as\[n = 2^{b}(2^{a-b+1}+1).\]
Lemma 1: All regular polygons with $n = 2^k + 1$ or $n=4$ have triangulations that meet the conditions.
By induction, it follows that we can cover all the desired $n$.
For $n = 3,4$, this is trivial. For $k>1$, we construct the diagonals of equal length $\overline{P_0P_{2^{k-1}}}$ and $\overline{P_{2^{k-1}+1}P_0}$. This partitions $\mathcal{P}$ into $3$ regions: an isosceles $\triangle P_0P_{2^{k-1}}P_{2^{k-1}+1}$, and two other regions. For these two regions, we can recursively construct the isosceles triangles defined above in the second paragraph. It follows that we have constructed $2(2^{k-1}-1) + (1) = 2^k-1 = n-2$ isosceles triangles with non-intersecting diagonals, as desired.\\
Lemma 2: If a regular polygon with $n$ sides has a working triangulation, then the regular polygon with $2n$ sides also has a triangulation that meets the conditions.
We construct the diagonals $\overline{P_0P_2},\ \overline{P_2P_4},\ \ldots \overline{P_{2n-2}P_0}$. This partitions $\mathcal{P}$ into $n$ isosceles triangles of the form $\triangle P_{2k}P_{2k+1}P_{2k+2}$, as well as a central regular polygon with $n$ sides. However, we know that there exists a triangulation for the $n$-sided polygon that yields $n-2$ isosceles triangles. Thus, we have created $(n) + (n-2) = 2n-2$ isosceles triangles with non-intersecting diagonals, as desired.
In summary, the answer is all $n$ that can be written in the form $2^{a+1} + 2^{b},\, a,b \ge 0$. Alternatively, this condition can be expressed as either $n=2^{k},\, k \ge 2$ (this is the case when $a+1 = b$) or $n$ is the sum of two distinct powers of $2$, where $1= 2^0$ is considered a power of $2$.\\
{\raggedright
\text{[Source]}:
}\\
USAMO 2008\\
</example math problem>\\

\#\# Summarization\\
The problem asks for the possible values of $n$ for a regular n-sided polygon that can be completely triangulated into isosceles triangles using non-intersecting diagonals. The solution involves analyzing the properties of the diagonals forming isosceles triangles and deducing that $n$ can be expressed in terms of powers of 2.
\\
\#\# Math domains\\
Mathematics -> Geometry -> Plane Geometry -> Polygons;\\

=== report over ===\\
\label{prompt:classify}
\end{tcolorbox}

\section{Prompts for theorem provers}
\subsection{Prompt for Vanilla Generation}
\vspace{0.5\baselineskip}
\begin{tcolorbox}[
    title=\textbf{\scriptsize Prompt for Vanilla Generation},
    colframe=c3,
    boxrule=0.5pt,
    colback=c4!5,
    fonttitle=\scriptsize, %
    fontupper=\scriptsize, %
    parskip=12pt
]
Complete the following Lean 4 code:\\
\`{}\`{}\`{}lean4\\
import Mathlib\\ \\
theorem omni\_theorem\_2669\\
$(x : \mathbb{Z}) \ (hx : x = 2018) : x^2 + 2 * x - x * (x + 1) = 2018 := by$
\label{textbox:prompt_vanilla}
\end{tcolorbox}

\vspace{2mm}
\subsection{Prompt for CoT Generation}
\begin{tcolorbox}[
    title=\textbf{\scriptsize Prompt for Cot Generation},
    colframe=c3,
    boxrule=0.5pt,
    colback=c4!5,
    fonttitle=\scriptsize, %
    fontupper=\scriptsize, %
    parskip=12pt
]
Complete the following Lean 4 code with explanatory comments preceding each line of code:\\
\`{}\`{}\`{}lean4\\
import Mathlib\\ \\
theorem omni\_theorem\_2669\\
$(x : \mathbb{Z}) \ (hx : x = 2018) : x^2 + 2 * x - x * (x + 1) = 2018 := by$
\label{textbox:prompt_cot}
\end{tcolorbox}

\vspace{2mm}

\subsection{Prompt for NL-Augmented CoT}
\begin{tcolorbox}[
    title=\textbf{\scriptsize Prompt for Cot with natural solution Generation},
    colframe=c3,
    boxrule=0.5pt,
    colback=c4!5,
    fonttitle=\scriptsize, %
    fontupper=\scriptsize, %
    parskip=12pt,
    breakable
]
Complete the following Lean 4 code with explanatory comments preceding each line of code:\\
\`{}\`{}\`{}lean4\\
import Mathlib\\
open Finset\\
theorem omni\_theorem\_4199 :\\
\(\exists n \in \mathbb{N}, \{s : \text{Finset} \mathbb{N} \mid \text{s.card} = 2017 \land \sum_{i \in s} i^2 = n \}. \text{ncard} \geq 2017 \text{:= by}\)\\
/-To determine if there exists a number \( n \) that can be expressed as the sum of 2017 perfect squares in at least 2017 distinct ways, we consider the properties and combinations of perfect squares.\\
\#\#\# Step 1: Understanding the Problem\\
The problem asks us to express a number \( n \) as the sum of 2017 perfect squares, \( n = a_1^2 + a_2^2 + \cdots + a_{2017}^2 \), where \( a_i \) are integers. Moreover, this can be done in at least 2017 different ways, meaning there are at least 2017 distinct sets of such integers.\\
\#\#\# Step 2: Exploring Perfect Squares\\
Perfect squares are non-negative numbers of the form \( k^2 \), where \( k \) is an integer. To construct different sums, we need to evaluate how the combinations of these squares can vary and still yield distinct sums that equate to the same \( n \).\\
\#\#\#  Step 3: Existence of Solutions\\
1. **Many Small Squares**: By choosing different arrangements of small perfect squares (like 0, 1, 4, 9, etc.), we can vary them freely since they don’t drastically alter the cumulative sum quickly. For instance, using 0 is trivial as it adds nothing to sums; including or excluding it in varying positions introduces variety.\\
2. **Adjusting a Larger Value**: Consider including a larger square, say \((k+1)^2\), and adjusting the rest of the terms accordingly. This diversity of combinations even with fixed values of \( a_i = 0 \) (i.e., not all contributing to sum) provides additional distinct setups.\\
\#\#\# Step 4: Conclusion\\
Given the vast number of combinations possible with 2017 variables, it is feasible to achieve at least 2017 distinct sums since:\\
- Choosing different subsets of minimal contributions (e.g., many zeros and small numbers) can still lead to varying sums.\\
- Incremental adjustments in a few selections using larger squares or varied middle-range integers allow differential assembly leading to the target sum.\\
Thus, there is indeed a number \( n \) that can be expressed as the sum of 2017 perfect squares in at least 2017 distinct ways.\\
Hence, the answer is:\\
\(\boxed{\text{Yes}}\)-/
\label{textbox:prompt_nat}
\end{tcolorbox}

\section{Prompt for Error Pattern Diagnosis}
\begin{tcolorbox}[
    title=\textbf{\scriptsize Prompt for Error Pattern Diagnosis},
    colframe=c3,
    boxrule=0.5pt,
    colback=c4!5,
    fonttitle=\scriptsize, %
    fontupper=\scriptsize, %
    parskip=12pt
]
**Role:** Lean4 Error Pattern Analyst\\

**Input:** You will be provided with a list containing 5 Lean4 code snippets. Assume these snippets contain errors or represent incorrect usage patterns.\\

**Task:** Analyze all 5 snippets and identify the **common features or error patterns** present across them.\\

**Output:** Generate a list of concise strings describing these common features. Each string should be a short label for the pattern.\\

**Constraints:**
* Focus *only* on identifying common features/errors across the provided 5 snippets.
* Do **not** correct or modify the code.
* Keep feature descriptions brief and informative (e.g., "Misuse of automated tactic", "Type mismatch in arguments", "Incorrect proof structure", "Syntax error in definition").\\

**Example Input Snippets (Conceptual):**
[Lean4 Code Snippet 1 (Incorrect), ..., Lean4 Code Snippet 5 (Incorrect)],\\

**Example Output:**
[
"Misuse of automated tactic": detailed reason, and exactly which problems (using problem id) make this fault.
....
]
each feature should be mutually exclusive, and the features should cover all the common features of the code.

**Analyze the following 5 Lean4 code snippets:**
\label{textbox:prompt_vfe}
\end{tcolorbox}

\section{Prompt for Error Pattern Categorization}
\begin{tcolorbox}[
    title=\textbf{\scriptsize Prompt for Lean4 Proof Error Classification},
    colframe=c3,
    boxrule=0.5pt,
    colback=c4!5,
    fonttitle=\scriptsize, %
    fontupper=\scriptsize, %
    parskip=12pt
]
**Role:** Lean4 Code Classifier

**Task:** Classify the given Lean4 code snippet into one or more of the following categories based on the identified error patterns:\\

1. Improper usage of the automation tactics\\
2. Incomplete or Placeholder Proof Steps\\
3. Misuse of rewriting/simplification tactics\\
4. Inadequate handling of inequalities\\
5. Redundant hypothesis introductions\\

**Output Format:**
Return a JSON object with the following structure:\\
\{\\
    "categories": ["category1", "category2", ...],\\
    "confidence": [0.8, 0.7, ...],  \# Confidence scores for each category\\
    "explanation": "Brief explanation of why these categories were chosen"\\
\}
\\
**Code to Classify:**
\label{textbox:prompt_classification}
\end{tcolorbox}

\end{CJK}
\end{document}